\documentclass[10pt,twocolumn,letterpaper]{article}

\usepackage{iccv}
\usepackage{epsfig}
\usepackage{graphicx}
\usepackage{chen}

\usepackage{booktabs}
\usepackage{colortbl}
\usepackage{times}  % DO NOT CHANGE THIS
\usepackage{helvet}  % DO NOT CHANGE THIS
\usepackage{courier}  % DO NOT CHANGE THIS
\usepackage{caption} % DO NOT CHANGE THIS AND DO NOT ADD ANY OPTIONS TO IT
\usepackage{textcomp}
\usepackage{stfloats}
\usepackage{url}
\usepackage{verbatim}
\usepackage{graphicx}
\usepackage{cite}
\usepackage{tabularx}
\usepackage{amsfonts,amssymb}
\usepackage{multirow}
\usepackage{bbding}
\usepackage{fontenc}
\usepackage{amsmath}

\usepackage[pagebackref,breaklinks,colorlinks]{hyperref}
\iccvfinalcopy % *** Uncomment this line for the final submission

\def\iccvPaperID{****} % *** Enter the ICCV Paper ID here

\ificcvfinal\fi

\begin{document}

%%%%%%%%% TITLE
\title{Sparse Sampling Transformer with Uncertainty-Driven Ranking for Unified Removal of Raindrops and Rain Streaks}

% \author{Sixiang Chen$^{\dag}$^{\inst{1}}\and
% Tian Ye$^{\dag}$^{\inst{2}}\and
% Jinbin Bai$^{\dag}$^{\inst{3}}\and
% Erkang Chen^{\inst{4}}\and
% Jun Shi^{\inst{5}}\and
% LeiZhu^{\inst{1}}\thanks{Corresponding author. $^{\dag}$Equal contribution.} \\
% $^1$School of Ocean Information Engineering, Jimei University, China \\
% $^2$College of Artificial Intelligence, Southwest University, China\\
% $^3$The Hong Kong University of Science and Technology (Guangzhou)\\
% }
%

\author{
Sixiang Chen$^{1,3*}$ 
\quad Tian Ye$^{{1,3}}$\thanks{Equal contributions.} 
\quad Jinbin Bai$^{{2}}$ \quad Erkang Chen$^{{3}}$ \\ \vspace{1mm} Jun Shi$^{{4}}$ \quad Lei Zhu$^{{1,5}}$\thanks{Lei Zhu (leizhu@ust.hk) is the corresponding author.}\\ \vspace{-0.5mm}
\small $^{1}$The Hong Kong University of Science and Technology (Guangzhou)\quad
\small $^{2}$National University of Singapore\quad \\ 
\small $^{3}$School of Ocean Information Engineering, Jimei University\\ 
\small $^{4}$Xinjiang University\quad
\small $^{5}$The Hong Kong University of Science and Technology\\
\small  $\left\{sixiangchen, owentianye \right\}$ \small @hkust-gz.edu.cn \\
\small jinbin.bai@u.nus.edu, ekchen@jmu.edu.cn, junshi2022@gmail.com, leizhu@ust.hk \\
{\small Project page: \url{https://ephemeral182.github.io/UDR_S2Former_deraining}}}

\maketitle
\ificcvfinal\thispagestyle{empty}\fi

\begin{abstract}
In the real world, image degradations caused by rain often exhibit a combination of rain streaks and raindrops, thereby increasing the challenges of recovering the underlying clean image. Note that the rain streaks and raindrops have diverse shapes, sizes, and locations in the captured image, and thus modeling the correlation relationship between irregular degradations caused by rain artifacts is a necessary prerequisite for image deraining. 
This paper aims to present an efficient and flexible mechanism to learn and model degradation relationships in a global view, thereby achieving a unified removal of intricate rain scenes. 
To do so, we propose a \underline{S}parse \underline{S}ampling Trans\underline{former} based on \underline{U}ncertainty-\underline{D}riven \underline{R}anking, dubbed $\textbf{UDR-S$^{2}$Former}$. 
Compared to previous methods, our UDR-S$^{2}$Former has three merits. First, it can adaptively sample relevant image degradation information to model underlying degradation relationships. 
Second, explicit application of the uncertainty-driven ranking strategy can facilitate the network to attend to degradation features and understand the reconstruction process. 
Finally, experimental results show that our UDR-S$^{2}$Former clearly outperforms state-of-the-art methods for all benchmarks. %with a low computational complexity and parameter amounts.
%Third, SnowFormer outshines advanced state-of-the-art desnowing networks and the prevalent universal image restoration transformers on six synthetic and real-world datasets. We will release our code and pre-trained model after the acceptance of our submitted manuscript.
\end{abstract}
\vspace{-0.6cm}
%%%%%%%%% BODY TEXT
\section{Introduction}
\vspace{-0.1cm}
Rain is a ubiquitous condition that negatively impacts various computer vision tasks~\cite{chang2020data,zhang2022dino}. In real-world rain scenes, raindrops and rain streaks are irregularly superimposed on clean images. Image deraining is employed to restore the clean images from the complex rain degradations. According to previous work~\cite{quaninonego}, the imaging model of precipitation, inclusive of rain streaks and raindrops, can be expressed as:

\begin{equation}
    \mathcal{R}_{d s}=(1-\mathcal{M}_r) \odot(\mathcal{B}+\mathcal{S})+\eta \mathcal{D},
\end{equation}
where $\mathcal{B}$ and $\mathcal{S}$ denote the clean background and rain streak map. $\mathcal{M}_r$ is a binary mask used to judge whether the pixel belongs to the raindrops or the background. $\mathcal{D}$ is the raindrops and $\eta$ means global atmospheric lighting coefficient. 

As noted by CCN~\cite{quaninonego}, removing rain streaks and raindrops in a unified manner cannot be achieved by simply combining separate methods for removing either. This is due to the complex nature of the physical models involved and the wide array of possible degradation combinations. Previous models developed to address singular forms of degradations~\cite{chen2023learning,prenet,DRD,quan2019deep,wang2019spatial,li2019heavy,yang2021recurrent} face notable obstacles when dealing with irregularly dispersed and diverse rain degradation types.

\begin{figure}[!t]
    \centering
    \setlength{\abovecaptionskip}{0.05cm} %调整caption与图的距离
    \setlength{\belowcaptionskip}{-0.6cm}%调整caption与下文的距离
    \includegraphics[width=8.5cm]{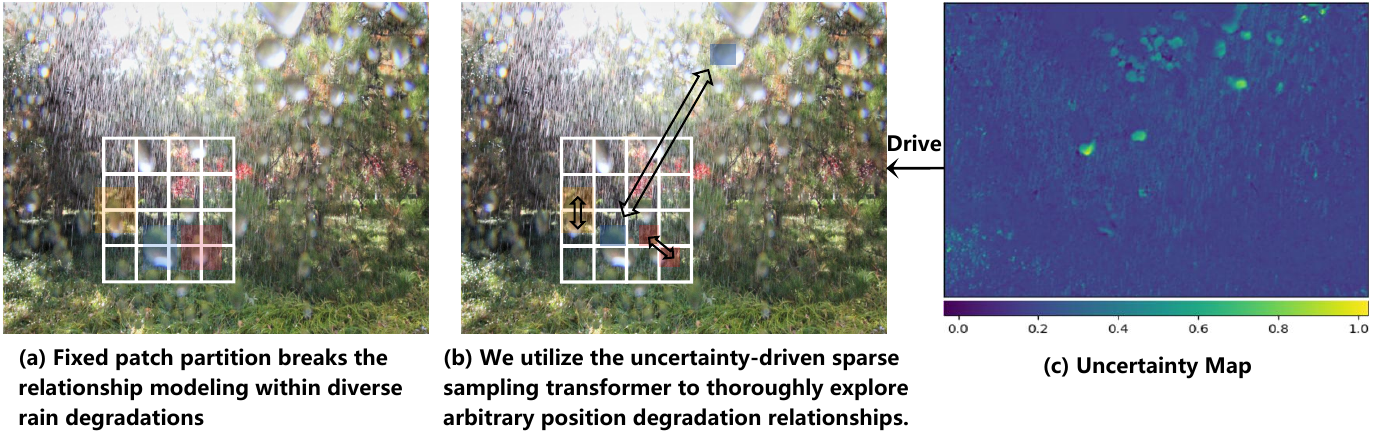}
    \caption{\small{Illustration of the breakdown of complex rain degradation relationships and the thumbnails of our main ideas. Colored places indicate degradations. Two-way arrows represents modeling between degradations.}}
    \label{banner}
\end{figure}

Specifically, current SOTA methods for image deraining primarily concentrate on using ViTs due to their abilities to model long-range dependencies~\cite{wang2022uformer,IDT,liang2022drt,jiang2022magic}. Among these methods, window-based self-attention~\cite{wang2022uformer,liang2022drt,IDT} has gained popularity due to its computational efficiency. 
However, as shown in Fig.\ref{banner},  we argue that utilizing window-based self-attention mechanisms can lead to incomplete degradation coverage, causing the breakdown of degradation relationships for unified rain degradation removal due to fixed window segmentation. This problem can be particularly pronounced when dealing with large raindrops or rain streaks at long distances simultaneously. However, in the case of complicated degradations, it is imperative to model the relationships between related forms of degradations.

\begin{figure*}[!t]
    \centering
    \setlength{\abovecaptionskip}{0.05cm} %调整caption与图的距离
    \setlength{\belowcaptionskip}{-0.6cm}%调整caption与下文的距离
    \includegraphics[width=16.8cm]{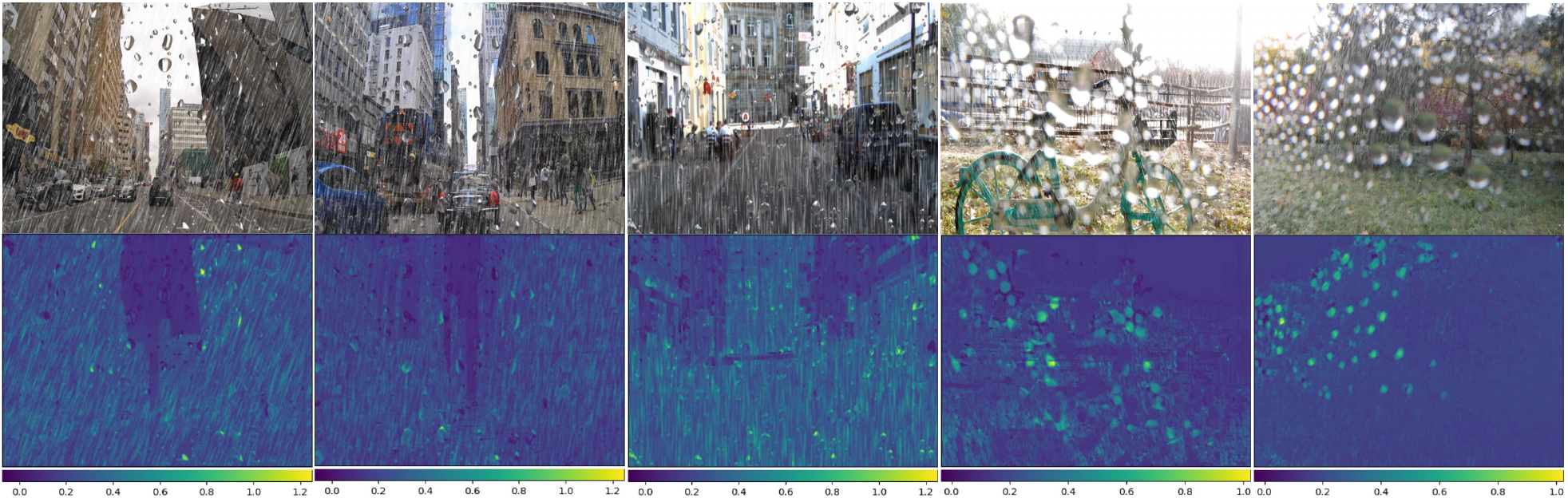}
    \caption{\small{\textbf{The uncertainty maps (bottom row) correspond to both real and synthetic samples (top row),} with more significant uncertainty appearing in areas with severe and complicated degradations. This observation motivates us to use uncertainty explicitly to represent knowledge about degradation and to improve the model's understanding of degradation restoration.}}
    \label{uncertainty}
\end{figure*}

Furthermore, with respect to the dense prediction task of rain removal, the density, shape, position, and size of raindrops and streaks are all uncertain, rendering it arduous for the network to restore clean images from diverse degradations. 
The CCN~\cite{quaninonego} requires using expensive and inflexible NAS to select an optimal architecture that effectively handles rain streaks and raindrops precisely.
Inspired by uncertainty modeling~\cite{kendall2017uncertainties} for image restoration, incorporating uncertainty learning can enhance the performance by reducing the error in model parameters~\cite{hong2022uncertainty}, or serve as a regular term constraint to enhance the prediction quality of regions characterized by high uncertainty~\cite{ning2021uncertainty,tong2022semi}.
Nevertheless, the mentioned design paradigm overlooks the importance of explicitly excavating uncertainty maps in facilitating the network's modeling of degradation features.
For intricate rain scenes, we claim that learning uncertainty estimation can in turn affect the network to better focus on the complicated rain degradation areas. As depicted in Fig.\ref{uncertainty}, the uncertainty map of the image exhibits greater concentration within the degraded region. It is our contention that by fully leveraging the properties of this uncertainty map, we can more effectively model the relationships of degradation and drive the network for the understanding of degradation restoration.

%-------------------------------------------------------------------------

Specifically, to address the problems above, we first design the Sparse Sampling Attention to deal with complicated rain scenes.
It sparsely learns the relevant degradation relationships from the entire image, thereby alleviating the drawback of large-scale degradation modeling in window-based attention. Concurrently, we leverage the uncertainty map to guide feature learning for capturing more discriminative sampling features. 
To be exact, in order to fully leverage uncertainty information to promote the sampling of degradation features, we propose a novel ranking strategy and present a Constraint Matrix based on it. Such design further restricts the degree of attention to various rain degradations in the sampling process, boosting the modeling of relationships between degradations. 
Moreover, when restoring partially degraded regions, we consider the internal difference of the uncertainty map and use the ranking of the Correlation Map to strengthen the network to restore the degraded area by leveraging clean cues within the local regions.

Overall, our contributions can be summarized as follows:
\begin{itemize}
\vspace{-0.2cm}
    \item An uncertainty-driven sparse sampling transformer that fully models the global degradation relationships in an efficient manner is proposed to remove diverse rain streaks and raindrops.
\vspace{-0.2cm}
    \item  We introduce a ranking strategy in the uncertainty map to enable the model to emphasize various rain degradation features in the sampling process through a constructed constraint matrix.
\vspace{-0.2cm}
    \item To enhance local reconstruction, we utilize the internal discrepancies within the uncertainty map to stimulate the network to extract credibly clean information. 

\end{itemize}
\vspace{-0.4cm}
\section{Related Works}
\vspace{-0.1cm}
\subsection{Single Image Deraining}
\vspace{-0.1cm}
\noindent\textbf{Rain streak removal.} Image restoration from adverse weather has made significant advancements over the years, owing to its paramount significance~\cite{chen2022msp,chen2022snowformer,ye2021perceiving,jin2023enhancing,liu2023nighthazeformer,DehazeYu,DehazeMMYu}. Recently, the field of single image deraining has been predominantly dominated by learning-based methods~\cite{JORDER,prenet,IDT,rescan,UMRL,mspfn,zhujoint,wang2019spatial,li2019heavy,yang2021recurrent,xu2021intensity,zhu2020learning}.
Zhu $et$ $al.$ ~\cite{zhujoint} proposed a joint optimization algorithm that involves iteratively removing rain streaks from the background layer and non-rain details from the rain layer. This is achieved through the incorporation of three essential priors. 
PreNet~\cite{prenet} offered a recurrent layer to leverage the inter-stage dependencies of deep features, thereby constructing the progressive recurrent network. 
UMRL~\cite{UMRL} proposed the uncertainty map to constrain the rain map in rainscapes and incorporate physical models to estimate the final deraining output. IDT~\cite{IDT} presented a transformer system comprising a complementary window-based transformer and spatial transformer, enabling improved capture of short- and long-range dependencies in rainy scenes.

\noindent\textbf{Raindrop removal.} Raindrops are frequently observed in rain scenes, and their diverse shapes and positions present difficulties removing them. Previous attempted to eliminate raindrops  utilizing various methods,~\cite{eigen2013restoring,qian2018attengan,quan2019deep,you2013adherent}. The Eigen $et$ $al.$~\cite{eigen2013restoring} firstly introduced the learning-based paradigm for raindrop removal. AttenGAN~\cite{qian2018attengan} adopted the combination of GAN and attention mechanism to recover the clean image from raindrop degradations. %Quan $et al.$~\cite{quan2019deep} leveraged shape-driven attention and channel re-calibration techniques to eliminate raindrops.

Most of the current design paradigms continue to create specialized networks for removing rain streaks or raindrops. A unified removal network design has still to be developed. While CCN~\cite{quaninonego} was the first to consider joint removal of rain degradation, their approach still requires expensive design such as NAS to address complex degradation characteristics. Other general-purpose networks~\cite{IDT,zamir2021restormer} disregard these two degradation characteristics, leading to high computation and parameters when improving performance.
\begin{figure*}[!t]
    \centering
    \setlength{\abovecaptionskip}{0.05cm} %调整caption与图的距离
    \setlength{\belowcaptionskip}{-0.6cm}%调整caption与下文的距离
    \includegraphics[width=16.5cm]{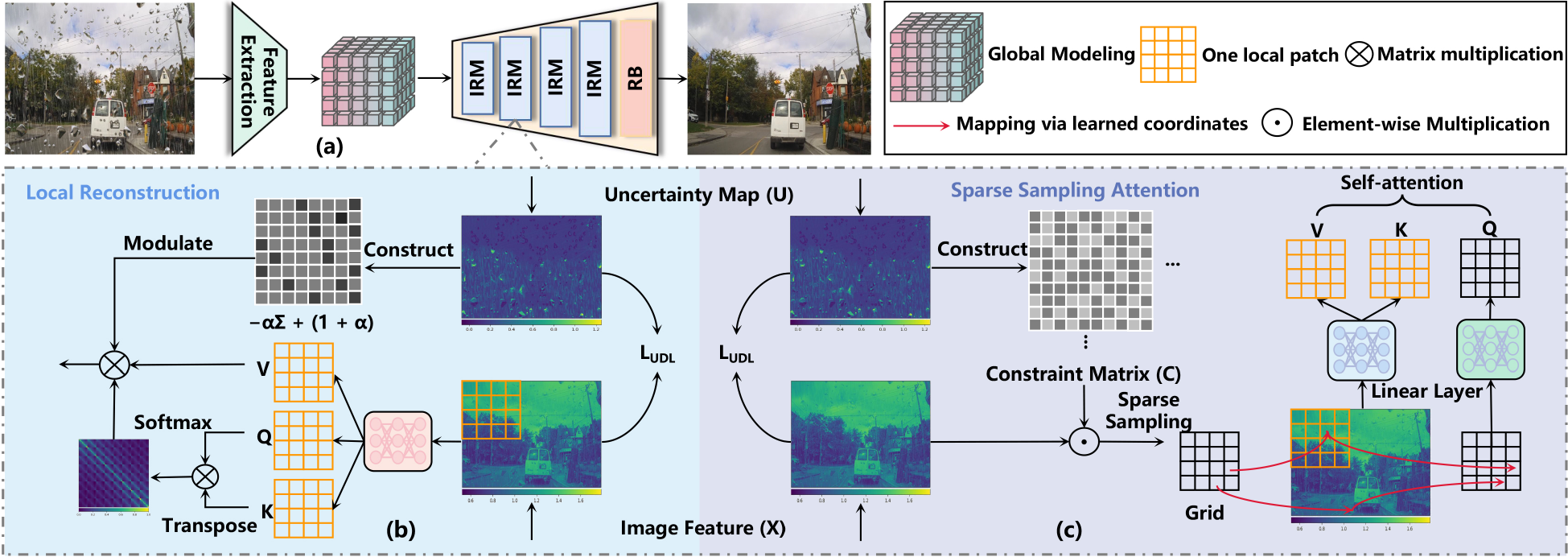}
    \caption{\small{The overview of our UDR-S$^{2}$Former pipline, which includes (a) our proposed restoration architecture, (b) the Local Reconstruction (LR) in the IRM module, (c) the Sparse Sampling Attention (SSA) constrained by the uncertainty map. For clarity, we only depict one local patch operation. The Feature Extraction and Global Modeling Modules, as well as the simple Refinement Block, are described in supplementary material due to page limitation.}}
    \label{overview}
\end{figure*}
\vspace{-0.2cm}
\subsection{Vision Transformers for Image Restoration}
\vspace{-0.1cm}
ViTs exhibited superior global modeling capabilities, resulting in impressive performance on low-level vision tasks~\cite{chen2022dual,zamir2021restormer,ART,chen2023dehrformer,song2022vision,Huang_2023_CVPR,valanarasu2022transweather,chen2023masked} compared with previous CNNs' paradigm~\cite{huang2022deep,chen2021attention,jin2021dc,jin2022unsupervised,ECLNet,jin2023estimating,NEURIPS2022_91a23b3e,Huang_2018_ECCV_Workshops,HPEU}. The window-based designs~\cite{liang2021swinir,wang2022uformer,IDT,liang2022drt,song2022vision} were widely employed to overcome the computational complexity issue of O($N^{2}$). Additionally, ART~\cite{ART} utilized a combination of sparse and dense self-attention to manage computational overhead and achieved SOTA outcomes. Restormer~\cite{zamir2021restormer} incorporated channel-based self-attention to circumvent the square-level complexity problem. Nevertheless, such designs limit the global receptive field of self-attention in spatial dimension and lack flexibility in dealing with intricate and changing degradations. We propose to use sparse sampling to address the aforementioned limitations by adaptively sampling information from the global field to meet the modeling requirements of the local.
\vspace{-0.2cm}
\subsection{Uncertainty in Deep Learning}
\vspace{-0.1cm}
According to Bayesian theory~\cite{kendall2017uncertainties}, uncertainty in deep learning can be classified into two types: (i) Aleatoric uncertainty, which refers to the inherent noise in the data. (ii) Epistemic uncertainty, which relates to the uncertainty of model parameters. Incorporating modeling uncertainty into the network can enhance its robustness and performance~\cite{badrinarayanan2017segnet,chang2020data}. In low-level fields, leveraging uncertainty can enable the network to prioritize reducing model prediction error~\cite{hong2022uncertainty} or serving as a loss function~\cite{ning2021uncertainty,tong2022semi,jin2022structure} to improve the reconstruction quality of areas with high uncertainty. Furthermore, uncertainty can assist in the more precise estimation of critical parameters in physical models~\cite{UMRL}. Same as ~\cite{ning2021uncertainty,tong2022semi}, we focus on aleatoric uncertainty in this paper. However, these paradigms do not explicitly utilize the distinctive characteristic of the uncertainty map to restrict the network's acquisition of features. This paper proposes leveraging uncertainty-driven ranking to facilitate the network to represent degradations.
\vspace{-0.3cm}
\section{Proposed UDR-S$^{2}$Former Pipline}
\vspace{-0.1cm}
\noindent\textbf{Network architecture.} The architecture of the proposed network is illustrated in Fig.\ref{overview}. The network accepts a rain image as input, conducts image processing within the network, and generates a high-quality restored image as output. Specifically, 1) the degraded image is fed into the Feature Extraction stage to acquire knowledge of features at various scales. This process is accomplished in four stages, each comprising basic convolutional blocks. 2) Upon completion of the feature extraction stage, we employ vanilla transformer to capture deep-level information and ensure the comprehensive utilization of global information\footnote{Restricted by the number of pages, we introduce the points 1) and 2) of our architecture in the supplementary material.}. 3) The Image Reconstruction Module (IRM) is employed to represent degradation relationships in the form of sparse sampling self-attention, which is driven by uncertain learning. Further, it can trigger clean cues excavation to guide the restoration for local reconstruction. Skip connections are utilized in both the feature extraction stage and each stage of the image reconstruction module.

\begin{figure*}[!t]
    \centering
    \setlength{\abovecaptionskip}{0.05cm} %调整caption与图的距离
    \setlength{\belowcaptionskip}{-0.6cm}%调整caption与下文的距离
    \includegraphics[width=16.8cm]{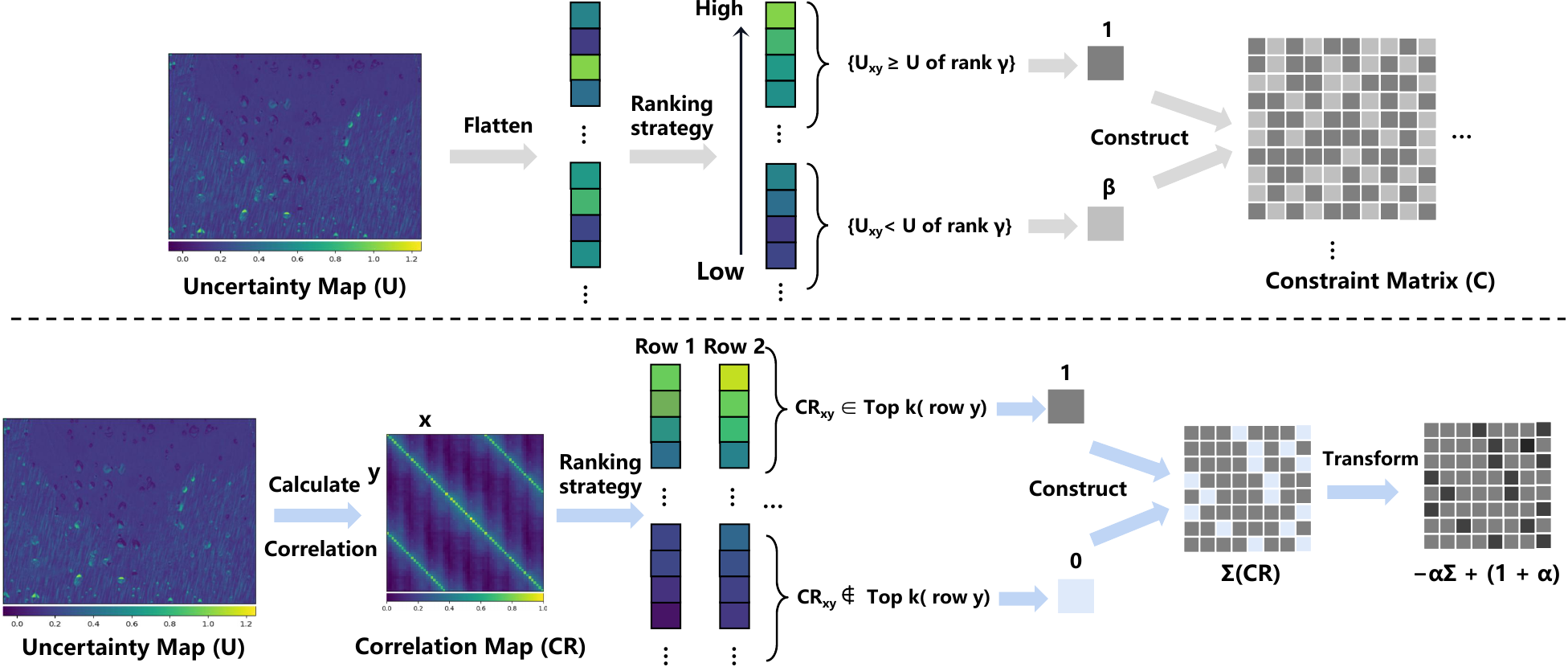}
    \caption{\small{\textbf{Top row}: Constructing a constraint matrix according to the ranking strategy derived from the uncertain map. To simplify the explanation, we only present a one-dimensional ranking strategy. \textbf{Bottom row}: The procedure for utilizing the correlation map and ranking strategy to produce the matrix utilized for modulating the self-attention map. We only present the correlation map and ranking strategy within a patch for easy visualization.}}
    \label{ranking}
\end{figure*}
\noindent\textbf{Preliminary for Uncertainty.} For image deraining, we model the difference between the image $\mathcal{B}_{gt}$ and its estimated deraining image $\hat{\mathcal{B}}$ as a Laplace distribution. The motivation for this choice is that the Laplace distribution is better suited for characterizing the edges of details in images than the Gaussian distribution. In addition, the L1 loss refers to the Laplace distribution~\cite{zhang2018ffdnet}. The likelihood function can be expressed as follows: 
\begin{equation}
    p(\mathcal{B}_{gt},\sigma;\mathcal{R}_{ds})=\frac{1}{2\sigma} \exp \left(-\frac{\lvert\lvert \hat{\mathcal{B}}-\mathcal{B}_{gt} \rvert\rvert_{1}}{\sigma}\right),
\end{equation}
where $\mathcal{B}_{gt}$ denotes the mean of this distribution. $\hat{\mathcal{B}}$ is output calculated by network from rain image $\mathcal{{R}}_{ds}$. $\sigma$ means the uncertainty (variancce) of deraining image $\hat{\mathcal{B}}$. To simplify the calculation, we transform the likelihood function into its log form and maximize it:
\begin{equation}
      \argmax_{\sigma} \ln p(\mathcal{B}_{gt},\sigma;\mathcal{R}_{ds})=(-\frac{\lvert\lvert \hat{\mathcal{B}}-\mathcal{B}_{gt} \rvert\rvert_{1}}{\sigma})-\ln\sigma.
\end{equation}
During the learning process, the aforementioned formula is inverted and utilized as a loss function, which is minimized for optimization. Additionally, we also follow ~\cite{ning2021uncertainty} to circumvent the issue of training instability resulting from the presence of uncertainties in the form of zero values.
Without being limited to a loss function, we aim to employ the ranking strategy to explicitly utilize estimated uncertainty. It enhances the modeling of complicated degradation relationships and drive the network to restore degraded regions.
\vspace{-0.6cm}
\subsection{Image Reconstruction Module}
\vspace{-0.1cm}
For the image reconstruction module of each stage, we can express it as:
\begin{equation}
    \text{IRM}(\mathcal{X},\mathcal{U}) = \left\{\text{SSA}(\mathcal{X},\mathcal{U}),\text{LR}(\mathcal{X},\mathcal{U})\right\},
\end{equation}
where $\left\{\cdot\right\}$ indicates that these two modules are alternately formed, with $\mathcal{X}$ denoting the image feature, and $\mathcal{U}$ representing the corresponding uncertainty map.
\vspace{-0.4cm}
\subsubsection{Sparse Sampling Attention Constrained by Uncertainty Map}\label{sec.ssa}
\vspace{-0.2cm}
The degradation of a large-scale area (raindrops) or long-distance span (rain streaks) can result in the loss of local relationships, particularly when a fixed window division is utilized. Moreover, a window-based design may limit the receptive field to a local area, leading to the loss of global knowledge when dealing with complicated degradations. However, it is crucial to fully exploit the modeling of degradation relationships for removing diverse rain degradations due to their complicated degraded properties.
%To effectively restore images with densely distributed rain marks and irregular raindrops, . 
In this part, we propose an uncertainty-driven sparse sampling approach to learn global coordinates for capturing associated rain degradations adaptively, effectively modeling degradations in diverse regions.
%Our approach employs an uncertainty ranking strategy to enhance the attention given to the degradation in the sparse sampling process.

For a given image feature ${\mathcal{X}^{\in \mathbb{R}^{ C\times H \times W}}}$, our primary objective is to establish a model that can accurately capture the degradation relationships. For each part of the degradations, the network should be prompted to concurrently consider correlation degradations occurring in other image regions to model it. Specifically, we let the network adaptively learn to match the degraded coordinates of other regions in the whole image, and map them to the same image patch, so as to model the degradation relationship at a low cost.
%We first partition feature ${\mathcal{X}^{\in \mathbb{R}^{ C\times H \times W}}}$ into $\mathcal{M}$ patches ${\mathcal{X}_{i}^{\in \mathbb{R}^{ C\times \frac{H}{\mathcal{M}} \times \frac{H}{\mathcal{M}}}}}$ .
The coordinates are learned as follows:
% \begin{equation}\label{coords}
%     \text{Coords}(x,y) = \mathcal{F}({\mathcal{X}^{\in \mathbb{R}^{ C\times  H\times W}}}),
% \end{equation}
\begin{figure}[!t]
    \centering
    \setlength{\abovecaptionskip}{0.05cm} %调整caption与图的距离
    \setlength{\belowcaptionskip}{-0.6cm}%调整caption与下文的距离
    \includegraphics[width=8.5cm]{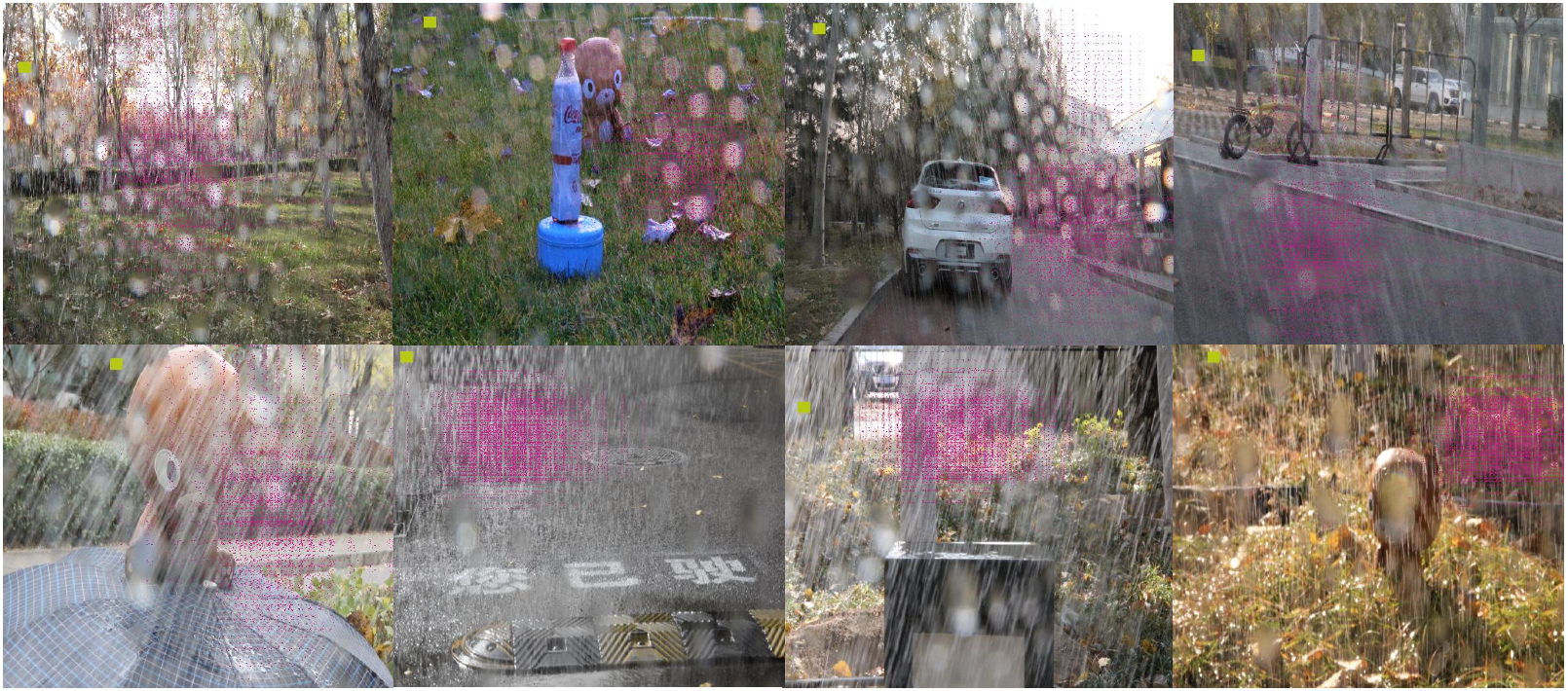}
    \caption{\small{\textbf{Visualization of sparse sampling.} The yellow region denotes an 8$\times$8 window patch. The red pixels are coordinates obtained from global sparse sampling to interact window patch for modeling the corresponding degradation relationship, which is not limited to the local patch (we draw multi-head sampling coordinates).}}
    \label{visualization}
\end{figure}
\begin{equation}\label{function}
   \mathcal{S}, \mathcal{B}  = \mathcal{F}({\mathcal{X}^{\in \mathbb{R}^{ C\times  H\times W}}}),
\end{equation}
where $\mathcal{F}(\cdot)$ represents the simple convolution and avgpooling operation to estimate scaling factors $\mathcal{S}^{\in \mathbb{R}^{ C\times  H\times W \times 2}}$ and biases $\mathcal{B}^{\in \mathbb{R}^{ C\times  H\times W \times 2}}$ (last dimension denotes x-axis and y-axis coordinates).
We use them to transform the original coordinates of the feature map by multiplication and addition:
\begin{equation}\label{coords}
    \text{Coords}^{T}(x,y) = \mathcal{G}(\underbrace{\text{Coords}^{O}(x,y)\times \mathcal{S} + \mathcal{B}\vphantom{\frac{2}{2}}}_{\text{\small Coordinate \, Transformation}}),
\end{equation}
% we use convolution and pooling layers to generate scaling factors $\mathcal{S}^{\in \mathbb{R}^{ C\times  H\times W \times 2}}$ and bias $\mathcal{B}^{\in \mathbb{R}^{ C\times  H\times W \times 2}}$ (last dimension denotes x-axis and y-axis coordinates), and use them to transform the original coordinates of the feature map by multiplication and addition. These transformed coordinates are used as grid parameters to sample the feature map globally, ensuring the degradation relationship modeling, as shown in the overview figure.
% where $\mathcal{F}(\cdot)$ represents the simple convolution and avgpooling operation to estimate 
scaling factors and biases change the sampling location for each patch. $\text{Coords}^{T}(x,y)$ denotes the coordinates of sparse sampling from the  global feature, which is leveraged to the local patch ${{\mathcal{X}_{i}^{S}}^{\in \mathbb{R}^{ C\times \frac{H}{\mathcal{M}} \times \frac{H}{\mathcal{M}}}}}$ ($i$ denotes the $i$-th patch of $\mathcal{M}^{2}$ patches) after carrying out the function $\mathcal{G}$ of $torch.nn.functional.grid\_sample$, as shown in Fig.\ref{overview} (c). We utilize the information obtained through global sparse sampling to match each degraded patches for corresponding degradation modeling and propose a novel attention mechanism, dubbed Sparse Sampling Attention (SSA):
\begin{equation}
\text{SSA}=\textbf{Softmax}\left(\frac{\mathbf{\mathcal{Q}}^{P}_{i} {{\mathcal{K}}^{S}_{i}}^{\text{T}}}{\sqrt{\mathcal{D}}}+p\right) \mathbf{\mathcal{V}}^{S}_{i},
\end{equation}
where ${\mathcal{Q}}^{P}_{i}$ denotes the queries projected from original feature of $i$-th patch.  ${\mathcal{K}}^{S}_{i}$ and ${\mathcal{V}}^{S}_{i}$ are obtained from global sampling feature. $\mathcal{D}$ means the dimension number and $p$ is the position embedding like \cite{swim}.

\noindent\textbf{Uncertainty map driven:} Drawing inspiration from the obvious representation of deteriorated regions in the uncertainty map, we aim to incorporate the uncertainty map as an explicit constraint to restrict the level of attention towards the degradations in the sparse sampling procedure, to ensure the modeling of the degradation relationships. The operation is presented in Fig.\ref{ranking}.

More specifically, with regard to the uncertain map $\mathcal{U}^{\in \mathbb{R}^{ C\times H \times W}}$, we employ a ranking strategy to construct a constraint matrix $\mathcal{C}^{\in \mathbb{R}^{ C\times H \times W}}$, which serves as a means of constraining the sampling process:
\begin{equation}
    \left[\mathcal{C}^{n}({\mathcal{U}^{n}})\right]_{x y}= \begin{cases}1 & \mathcal{U}^{n}_{x y}\geq \mathcal{U}^{n} \:\text{of rank} \:\gamma  \\ \beta & \text { otherwise }\end{cases},
\end{equation}
where $\beta$ and $\gamma$ are the constraint factor and a threshold value of $n$-th dimension. Such a constraint matrix contains crucial cues from the uncertainty map with almost no computational overhead, and we utilize it to regularize our modeling of degradation relationships. The Eq.\ref{function} is further optimized as:
\begin{equation}
   \mathcal{S}^{U}, \mathcal{B}^{U}  = \mathcal{F}({\mathcal{X}^{\in \mathbb{R}^{ C\times  H\times W}}\times \mathcal{C}_{x y}}),
\end{equation}
% \begin{equation}
%         \text{Coords}(x,y) = \mathcal{F}({\mathcal{X}^{\in \mathbb{R}^{ C\times  H\times W}}}* \mathcal{C}_{x y}),
% \end{equation}
wherein $\times$ represents the element-wise multiplication. The $\beta$ of $\mathcal{C}_{x y}$ reduces the influence of irrelevant background areas in multiplication form. $\mathcal{S}^{U}$ and $\mathcal{B}^{U}$ are leveraged to obtain robust coordinates via Eq.\ref{coords}, which promotes the global sparse sampling while mitigating the interference of the background region. Thereby it facilitates the network to concentrate on degradation relationship modeling.
% This operation may be construed as implicitly mitigating the influence of the background region on sparse sampling, thereby facilitating the network to concentrate on degradation relationship modeling.

Our visualization is depicted in Fig.\ref{visualization}. As we expected, the sampled red points adaptively have related degradation situations with the target point (see 8$\times$8 yellow window). The sparse sampling strategy with uncertainty helps model the relation between local and long-range correlation degradations. Additionally,
The most concentration of sampled pixels in a larger area maintains some semantic information and coherence, despite being sampled at a distance.

\noindent\textbf{\textcolor[RGB]{178, 105, 87}{Discussion I}: The merits of our sparse sampling compared with previous related work for image restoration.} Compared to the latest sparse attention~\cite{ART},  which permitted each token to interact with a limited number of tokens, and with a fixed interval size. However, it still involved a manually designed mechanism. In complex rainy scenes, such an approach cannot flexibly capture degradation information from different locations. Our design aims to enable the network to learn the coordinates that can sample degradation information from any position, leading to an enhanced modeling performance of the degradation relationships and network flexibility.

\noindent\textbf{\textcolor[RGB]{63, 167, 246}{Discussion II}: uncertainty driven compared with prior paradigm and rain mask.} Regarding uncertainty, the previous UMRL~\cite{UMRL} approach used the uncertainty to optimize the rain streak map based on the physical model of the rain streak. Still, it only focused on optimizing the final map, ignoring the network's learning of the intermediate features. Our method uses uncertainty explicitly to drive the network to provide motivation-driven constraints and optimization for the degradation modeling and local reconstruction processes.
Compared to the learned rain mask needed GT~\cite{purohit2021spatially}, \textbf{i)} uncertainty map has a theoretical unsupervised loss for practicing. \textbf{ii)}. It outperforms the rain mask in representing the network's focus on intricate rain areas beyond mere location. \textbf{iii)}. For large-area raindrops with a complicated physical model, it is difficult to learn a rain mask, but the uncertainty map can solve it well by using the discriminative representation of degradations.
\begin{figure}[!t]
    \centering
    \setlength{\abovecaptionskip}{0.05cm} %调整caption与图的距离
    \setlength{\belowcaptionskip}{-0.5cm}%调整caption与下文的距离
    \includegraphics[width=8.5cm]{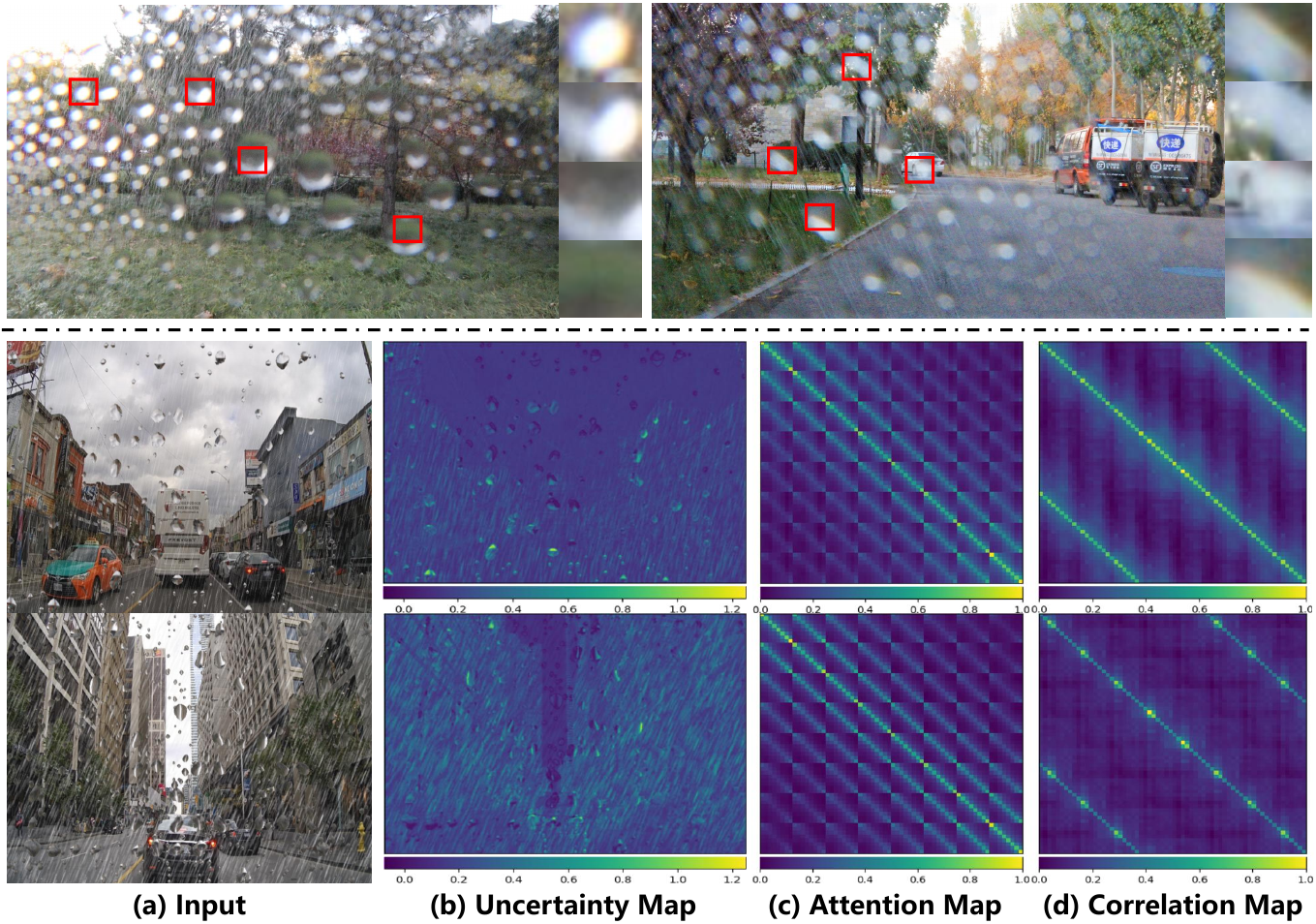}
    \caption{\small{\textbf{Top row}: Our motivation about needing to strengthen the network to exploit non-degradation regions. Most information is lost due to the complicated rain degradations in the real world, so degradation-free information is needed to reconstruct the clean area. \textbf{Bottom row}: Visualizations on uncertainty map, correlation map and self-attention map. The attention points of the correlation map and the self-attention map differ. We can utilize the correlation map to direct the network's attention towards areas with significant differences, thus enhancing the restoration process through self-attention.}}
    \label{motivation_local}
\end{figure}
\vspace{-0.4cm}
\subsubsection{Local Reconstruction with Correlation Ranking}\label{sec.lr}
\vspace{-0.1cm}
In rain scenes, information in locally degraded regions is often obscured by rain streaks or raindrops, making it difficult to restore, see Fig.\ref{motivation_local}. To address this issue, it is vital to leverage clean cues in the rain-free area. Building on the discriminative representation of degraded areas and other regions in the uncertainty map, we propose using the uncertainty map to generate a correlation map. This map explicitly adjusts the attention map through a ranking approach, as shown in Fig.\ref{ranking} and Fig.\ref{overview}(b), which promotes the network to better leverage clean cues for reconstruction in the form of Query-Key.
    
We begin by partitioning uncertainty map $\mathcal{U}^{\in \mathbb{R}^{ C\times H\times W}}$ into local patches ${\mathcal{U}^{P}}^{\in \mathbb{R}^{ C\times \frac{H}{\mathcal{M}} \times \frac{H}{\mathcal{M}}}}$ that correspond to those patches in the feature map. The presence of discriminative disparities within each patch of the uncertainty map motivates us to calculate the correlation map $(\mathcal{CR})$ between them:
\begin{equation}
    \mathcal{CR}_{i} = \mathcal{U}_{i}^{P}\times{\mathcal{U}_{i}^{P}}^{\text{T}}.
\end{equation}
Upon obtaining the correlation map, our purpose is to leverage the degradation-free region to facilitate the restoration process of the degraded areas during image reconstruction. Significant differences (low correlation) between degradations and background in the uncertainty map can be considered a prompt, promoting the network to use clean cues from the background for local reconstruction.
To this end, we modulate the self-attention map by selecting the low-correlation regions in the correlation map:
\begin{equation}
\text{LR}=\textbf{Softmax}\left(\frac{\mathbf{\mathcal{Q}}^{{P}}_{i} {{\mathcal{K}}^{{P}}_{i}}^{\text{T}}}{\sqrt{\mathcal{D}}}\times\underbrace{[-\alpha \boldsymbol{\Sigma}_{i}+(1+\alpha) ]+p\vphantom{\frac{2}{2}}}_{\text{\small Modulation \, Processing}}\right)\mathbf{\mathcal{V}}^{{P}}_{i},
\end{equation}
where $\alpha$ is a modulation factor. The $\Sigma$ is acquired via the \text{Top k} ranking approach, which can be expressed as follows:
\begin{equation}
    \left[\boldsymbol{\Sigma}_{i}({\mathcal{CR}_{i}})\right]_{x y}= \begin{cases}1 & (\mathcal{CR}_{i})_{x y} \in \operatorname{Top}\mathrm{k}(\text { row } y) \\ 0 & \text { otherwise }\end{cases}.
\end{equation}
In the experiments section, the key parameters of the above used are further studied. About our image reconstruction module, the self-attention FFNs are identical to vanilla vision transformer~\cite{vit}.

% \begin{equation}
%     \mathcal{S} = \mathcal{U}\times\mathcal{U}^{\text{T}}
% \end{equation}

\vspace{-0.3cm}
\section{Loss Function}
\vspace{-0.1cm}
Our UDR-S$^{2}$Former network will predict the derained result and an uncertainty map. Hence, the total loss of our network is defined as follows:
\begin{equation}
\begin{aligned}
    \mathcal{L}_{total} &= \lambda_{1} \mathcal{L}_{psnr} (\mathcal{B}_{pre}, \mathcal{B}_{gt}) + \lambda_{2} \mathcal{L}_{perceptual}(\mathcal{B}_{pre}, \mathcal{B}_{gt})\\
    &+\lambda_{3} \mathcal{L}_{UDL}(\mathcal{B}_{pre}, \mathcal{U}), 
\end{aligned}
\end{equation}
where $\mathcal{B}_{pre}$ and $\mathcal{B}_{gt}$ denote the predicted result of image deraining and the corresponding ground truth.
$\mathcal{U}$ denotes the predicted uncertainty map.
$\mathcal{L}_{psnr}$ and $\mathcal{L}_{perceptual}$ are the PSNR and perceptual losses, see~\cite{chen2022dual,chen2021hinet}.
$\mathcal{L}_{UDL}$ denotes the uncertainty loss, see ~\cite{ning2021uncertainty,tong2022semi} for its definition.
$\lambda_{1}$, $\lambda_{2}$ and $\lambda_{3}$ are empirically set to 1, 0.2 and 1.

\if 0
We first follow~\cite{chen2022dual} to utilize the PSNR loss~\cite{chen2021hinet} and the perceptual loss to compute the difference between the predicted deraining of in our UDR-S$^{2}$Former network and the underlying ground truth. 
%%

Then, we the same uncertainty loss~\cite{ning2021uncertainty,tong2022semi} to compute the loss of our uncertainty map prediction. 
Moreover, to supervise uncertainty maps, we adopt the same uncertainty loss~\cite{ning2021uncertainty,tong2022semi}:
\begin{equation}
      \mathcal{L}_{UDL}=(\frac{\lvert\lvert \hat{\mathcal{B}}-\mathcal{B}_{gt} .\rvert\rvert_{1}}{\sigma})+2\ln\sigma.
\end{equation}
Hence, the total loss $\mathcal{L}_{total}$ of our network is computed by:
\begin{equation}
    \mathcal{L}_{total} = \lambda_{1} \mathcal{L}_{psnr} (\mathcal{B}_{pre}, \mathcal{B}_{gt}) + \lambda_{2} \mathcal{L}_{perceptual}(\mathcal{B}_{pre}, \mathcal{B}_{gt})+\lambda_{3} \mathcal{L}_{UDL}(\mathcal{B}-\mathcal{B}_{gt}), 
\end{equation}
where weights $\lambda_{1}$, $\lambda_{2}$ and $\lambda_{3}$ are empirically set to 1, 0.2 and 1. 
\fi

\vspace{-0.4cm}
\section{Experiments}
\vspace{-0.1cm}
\subsection{Implementaion Details}
\vspace{-0.1cm}
Our UDR-S$^{2}$Former model utilizes a 5-level encoder-decoder architecture. Within this architecture, the number of channel dimensions increases to $\left\{16, 32, 64, 128, 256\right\}$ across levels 1 to 5. Additionally, the Feature Extraction stage contains $\left\{4, 6, 7, 8\right\}$ convolutional blocks, while the transformer block number in the latent layer is 8, with 16 heads. Regarding reconstruction, we employ $\left\{3, 6, 7, 8\right\}$ Image Reconstruction Modules, composed alternately of Sparse Sampling Attention and Local Reconstruction, with $\left\{1, 2, 4, 8\right\}$ heads. The window size is fixed at 8$\times$8. Finally, we set the constraint factor $\beta$, threshold value $\gamma$, modulation factor $\alpha$, and the k value for the ranking strategy to 0.6, 0.8, 0.2, and 0.8, respectively. For the self-attention mechanism in this paper, we all use multi-head self-attention, which is consistent with vanilla ViT~\cite{vit}. Additionally, during the last image reconstruction module of each stage, we convert the output feature and corresponding uncertainty map into the final image format, and the $\mathcal{L}_{UDL}$ is utilized to supervise the learning process of uncertainty map.
%In the Attention Refinement Head, we employ six refinement blocks for handling each stage of features with the help of degradation-aware position encoding.

During the training phase, we employed the Adam optimizer with initial momentum $\beta_{1}= 0.9$ and $\beta_{1}= 0.999$. We initially set the learning rate to 0.0003 and utilized a cyclic learning rate adjustment strategy, whereby the maximum learning rate is set to 0.00036.  We trained our model with a data augmentation strategy that included randomly cropping 256 × 256 patches and using horizontal flipping and random image rotation to a fixed angle. Our training process involved $6\times10^{5}$ steps. To leverage perceptual loss, we used the first and third layers of VGG19~\cite{simonyan2014very}. Our model is implemented using PyTorch~\cite{paszke2019pytorch} and the RTX 3090 GPU.
\begin{table}[!t]
\setlength{\abovecaptionskip}{0cm} %调整caption与图的距离
\setlength{\belowcaptionskip}{-0.4cm}%调整caption与下文的距离
\centering
\caption{\small{Quantitative results compared with SOTA methods on the Rain200H~\cite{JORDER} and Rain200L~\cite{JORDER} datasets. Underline and bold indicate the first and second best results.}}\label{rainstreaksresults}
\resizebox{6cm}{!}{
\renewcommand\arraystretch{1.1}

\begin{tabular}{c|cc|cc}
\hline\thickhline
\rowcolor{mygray}
& \multicolumn{2}{c}{Rain200H~\cite{JORDER}} & \multicolumn{2}{|c}{Rain200L~\cite{JORDER}}\\\cline{2-5}
 \rowcolor{mygray}
\multirow{-2}*{Method} & PSNR $\uparrow$ & SSIM $\uparrow$ &PSNR $\uparrow$ & SSIM $\uparrow$\\\cline{1-5}
GMM~\cite{GMM}  & 14.71 & 0.430 & 28.99 &0.875\\
JCAS~\cite{JCAS}& 14.87 & 0.471 & 30.05 &0.897 \\
DDN~\cite{DDN} & 26.36 & 0.803 & 34.93 & 0.958\\
NLEDN~\cite{NLEDN}  & 29.51 & 0.891 & 38.56&0.980\\
RESCAN~\cite{rescan}  & 27.45 & 0.821 & 35.08 & 0.959\\
PreNet~\cite{prenet}  & 29.04 & 0.890 & 37.12 & 0.976\\
UMRL~\cite{UMRL}  & 28.71 & 0.887 & 36.43 & 0.973\\
JORDER-E~\cite{JORDER-E}  & 28.58 & 0.876&36.90 &0.973  \\ 
MSPFN~\cite{mspfn}  & 29.66 &{0.890} &39.48 &0.984\\
CCN~\cite{quaninonego}  & 29.99 & 0.914 & 38.26 &0.981 \\
MPRNet~\cite{mpr} & 30.76 & 0.908 & 39.89  & 0.985 \\
 DGUNet~\cite{mou2022deep} &${30.85}$ &${0.911}$ &${40.23}$  &${0.986}$ \\
 Uformer~\cite{wang2022uformer} & 30.80 & 0.911 & 40.20  & 0.986  \\
 Restormer~\cite{zamir2021restormer} & 31.39 & 0.916 & 40.58  & 0.987\\
 IDT~\cite{IDT} &{\underline{32.10}} &\underline{0.934} &\underline{40.74}  &\underline{0.988} \\
NAFNet~\cite{chen2022simple} &${30.98}$ &${0.912}$ &${{40.45}}$  &{0.987}  \\
\hline\hline  UDR-S$^{2}$Former& {\bbetter{25}{12}{\textbf{32.59}}{0.49}} & {\bbetter{25}{12}{\textbf{0.937}}{0.003}} & {\bbetter{25}{12}{\textbf{40.96}}{0.22}}&{\bbetter{25}{12}{\textbf{0.989}}{0.001}} \\
\hline\thickhline
\end{tabular}}
\vspace{-0.1cm}
\end{table}
\vspace{-0.6cm}
\subsection{Evaluation Metrics and Datasets}
\vspace{-0.2cm}
\noindent\textbf{Evaluation Metrics.} In accordance with prior deraining methodologies~\cite{quaninonego,IDT}, follow~\cite{gu2020pipal,gu2020image}, the performance of the model is evaluated using PSNR~\cite{PSNR} and SSIM~\cite{SSIM}. As suggested in ~\cite{JORDER,quaninonego}, the assessment of PSNR and SSIM is based on the luminance channel, which refers to the Y channel of the YCbCr space.
%, the same as~\cite{JORDER,quaninonego}.

\noindent\textbf{Rain Streak Datasets.} In order to evaluate the effectiveness of our rain streak removal method, we follow the strategy of previous work~\cite{quaninonego} and select two benchmark datasets, namely Rain200H and Rain200L~\cite{JORDER}. Both two datasets have 1800 synthetic images for training and 200 images for testing.

\noindent\textbf{Raindrop Datasets.} We also evaluate our method in a raindrop dataset (i.e., AGAN-Data) collected by Qian $et$ $al.$~\cite{qian2018attengan}. AGAN-Data has 861 images for training and 58 images for testing. 
\begin{figure*}[!t]
    \centering
    \setlength{\abovecaptionskip}{0.03cm} %调整caption与图的距离
    \setlength{\belowcaptionskip}{-0.3cm}%调整caption与下文的距离
    \includegraphics[width=17.2cm]{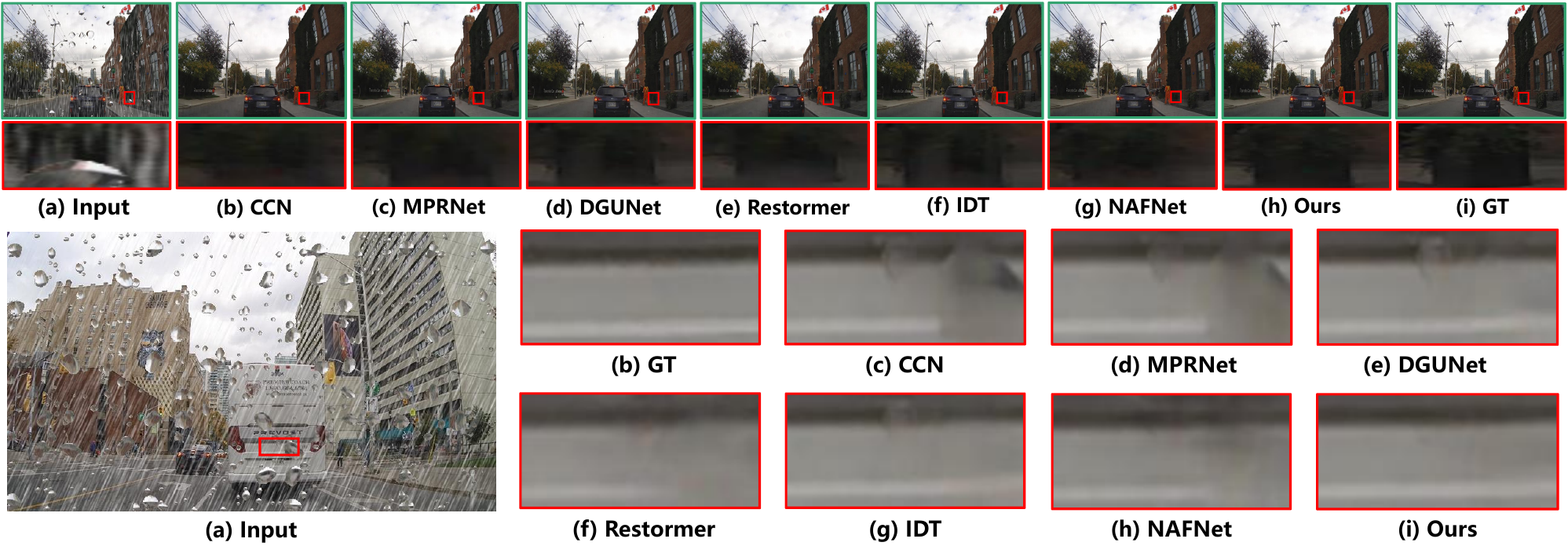}
    \caption{\small{Visual comparisons of removal for raindrops and rain streaks from RainDS-Syn~\cite{quaninonego} dataset. Zoom it for a better illustration.}}
    \label{syn1}
\end{figure*}

 \begin{table*}[!t]
\centering
\setlength{\abovecaptionskip}{0.1cm} %调整caption与图的距离
\setlength{\belowcaptionskip}{-0.0cm}%调整caption与下文的距离
\caption{\small{Quantitative comparisons of various SOTA approaches on the RainDS~\cite{quaninonego} benchmark. Our proposed method outperforms all other approaches on both synthetic and real-world datasets, including all types of precipitation (i.e., rain streaks (RS), raindrops (RD), and a combination of both (RDS)). The results are indicated in bold and underline, representing the first and second-best performances, respectively. $\uparrow$ means the higher is better. The \#GFLOPs are calculated by 256$\times$256 image resolution for a fair comparison.}}\label{raindsresults}
\resizebox{17.2cm}{!}{
\setlength\tabcolsep{2pt}
\renewcommand\arraystretch{1.1}
\begin{tabular}{c|c|cccccc|cccccc|cccc}
%\toprule[1.4pt]
\hline\thickhline
\rowcolor{mygray}
  & &\multicolumn{6}{c|}{ \textbf{RainDS-Syn} } &\multicolumn{6}{c|}{ \textbf{RainDS-Real} } &   & \\\cline{3-14}
   \rowcolor{mygray}
   & & \multicolumn{2}{c|}{ \textbf{RS} }& \multicolumn{2}{c|}{ \textbf{RD} } & \multicolumn{2}{c|}{ \textbf{RDS} } & \multicolumn{2}{c|}{ \textbf{RS} }& \multicolumn{2}{c|}{ \textbf{RD} } & \multicolumn{2}{c|}{ \textbf{RDS} }  &  &  \\\cline{3-14}
 \rowcolor{mygray}
\multirow{-3}*{Method}  &\multirow{-3}*{Venue} & PSNR $\uparrow$ & SSIM $\uparrow$ & PSNR $\uparrow$ & SSIM $\uparrow$ & PSNR $\uparrow$ & SSIM $\uparrow$ & PSNR $\uparrow$ & SSIM $\uparrow$ & PSNR $\uparrow$ & SSIM $\uparrow$ & PSNR $\uparrow$ & SSIM $\uparrow$ & \multirow{-3}*{\textbf{\#Param}} & \multirow{-3}*{\textbf{\#GFLOPs}} \\\hline\hline
GMM~\cite{GMM} & \textit{CVPR'2016} & 26.66 & 0.781 &23.04 &0.793& 21.50 & 0.669 & 23.73&0.560 & 18.60&0.554&21.35&0.576&-&-\\
JCAS~\cite{JCAS}& \textit{ICCV'2017} &26.46 & 0.786 &23.15 &0.811& 20.91 & 0.671 & 24.04&0.556& 18.18&0.555 & 21.22&0.585 &-&- \\
DDN~\cite{DDN} & \textit{CVPR'2017}& 30.41 &0.869 & 27.92 & 0.885 & 26.85 & 0.796 & 24.85&0.683& 23.12&0.642&22.47&0.606&-&-\\
NLEDN~\cite{NLEDN} &  \textit{ACMMM'2018} & $36.24$ & $0.958$ & 34.87&0.957& 32.13&0.917 & 27.02 & 0.723& 24.71 & 0.671 &24.06&0.650&-&-\\
RESCAN~\cite{rescan} & \textit{ECCV'2018} & {30.99} &{0.887} &{29.90} &{0.907} & {27.43} &{0.818}  & 26.70&0.683& 24.23&0.637&23.23&0.587 &0.15M&32.32G\\
PreNet~\cite{prenet} &  \textit{CVPR'2019} & $36.63$ & $0.968$ &34.58 &0.964& 32.21 & 0.934 & 26.43&0.729& 24.42&0.679&23.57&0.649&0.17M& 66.58G\\
UMRL~\cite{UMRL} & \textit{CVPR'2019} & 35.76 &{0.962} & 33.59 &0.958& 31.57&0.929 & 25.89&0.726& 23.93&0.676&23.01&0.647&0.98M&16.50G \\
JORDER-E~\cite{JORDER-E} &  \textit{TPAMI'2019} & 33.65 & 0.925&33.51 &0.944& 30.05&0.870 & 26.56&0.713& 24.34&0.662&23.54&0.629&4.17M & 273.68G  \\ 
MSPFN~\cite{mspfn} & \textit{CVPR'2020} & 38.61 &{0.975} &36.93 &0.973& 34.08&0.947 & 26.45&0.727& 24.49&0.681&24.11&0.651&21.00M&708.44G \\
CCN~\cite{quaninonego} & \textit{CVPR'2021} & 39.17 &{0.981} &37.30 &0.976& 34.79&0.957 & 27.46&0.737& 25.14&0.701&24.93&0.679&3.75M&245.85G \\
 % TransWeather~\cite{valanarasu2022transweather} & \textit{CVPR'2022} &${31.76}$ &${0.93}$ &${28.29}$  &${0.92}$ & ${31.82}$ &  ${0.93}$& 32.34&0.95& 31.57&0.94 &21.90M&5.64G\\
 % TKL~\cite{chen2022learning} & \textit{CVPR'2022} & $33.89$ & $0.96$ &30.82 &0.96& 34.37&0.95 & 35.67&0.97& 34.17&0.96&31.35M&41.58G \\\hline
 MPRNet~\cite{mpr} & \textit{CVPR'2021}&{40.81} &${0.981}$ &${37.03}$  &${0.972}$ & ${34.99}$ &  ${0.956}$& 27.29&0.736& 25.26&0.701 &24.96&0.681&3.64M&148.55G\\
 %SWinIR~\cite{liang2021swinir} & \textit{ICCVW'2021}&{40.67} &{0.974} &{38.14}  &${0.968}$ & 35.03 & ${0.952}$ & 27.53&0.743& 25.69&0.701&25.05&0.681&26.10M&140.99G \\
 DGUNet~\cite{mou2022deep} & \textit{CVPR'2022}&${41.09}$ &${0.983}$ &${37.56}$  &${0.975}$ & ${35.34}$ &  ${0.959}$& 27.52&0.737& 25.33&0.702&24.99&0.683&12.18M&199.74G\\
 Uformer~\cite{wang2022uformer} & \textit{CVPR'2022}&${40.69}$ &${0.972}$ &${37.08}$  &${0.966}$ & ${34.99}$ & ${0.954}$& 26.89&0.730& 25.31&0.701&24.83&0.686&20.63M&43.86G \\

 Restormer~\cite{zamir2021restormer} & \textit{CVPR'2022(Oral)}&{{41.42}} &{0.980} &{38.78}  &${0.976}$ & {36.08} & ${0.961}$ & 27.39&0.742& 25.38&0.702&24.92&0.685&26.10M&140.99G \\
% &MAXIM~\cite{tu2022maxim} &\textit{CVPR'2022(Oral)}&${31.76}$ &${0.93}$ &${28.29}$  &${0.92}$ & ${31.82}$ &  ${0.93}$& -&-& -&- \\
 IDT~\cite{IDT} & \textit{TPAMI'2022}&{\underline{41.61}} &\underline{0.983} &\underline{39.09}  &\underline{0.980} & \underline{36.23} & ${0.960}$ & \underline{27.51}&\underline{0.743}& \underline{25.67}&\underline{0.706}&\underline{24.99}&\underline{0.689}&16.00M&61.90G \\
NAFNet~\cite{chen2022simple} &\textit{ECCV'2022}&${40.39}$ &${0.972}$ &${{37.23}}$  &{0.974}  &${34.99}$ &{0.957} & {27.49} & {0.729} & {25.23}&{0.701}&24.64&0.663&40.60M&16.19G \\\hline\hline
UDR-S$^{2}$Former (Ours) & \textit{ICCV'2023} &{\bbetter{25}{15}{\textbf{42.39}}{0.78}}&{\bbetter{25}{15}{\textbf{0.988}}{0.005}} & {\bbetter{25}{15}{\textbf{39.78}}{0.69}} &{\bbetter{25}{15}{\textbf{0.983}}{0.003}}& {\bbetter{25}{12}{\textbf{36.91}}{0.68}} &{\bbetter{25}{12}{\textbf{0.966}}{0.006}} & {\bbetter{25}{12}{\textbf{27.90}}{0.39}} & {\bbetter{25}{12}{\textbf{0.745}}{0.002}} & {\bbetter{25}{12}{\textbf{26.01}}{0.34}}&{\bbetter{25}{12}{\textbf{0.709}}{0.003}}&{\bbetter{25}{12}{\textbf{25.52}}{0.53}}&{\bbetter{25}{12}{\textbf{0.691}}{0.002}}&8.53M&21.58G \\

\hline\thickhline
\end{tabular}}
\end{table*}
\noindent\textbf{Raindrops and Rain Streak Datasets.} 
Our paper aims to develop a methodology that effectively addresses the challenges posed by the presence of large-area raindrops and complex rain streaks. To evaluate the proposed approach, we utilize the RainDS benchmark dataset~\cite{quaninonego}, which includes both real-world and synthetic images, dubbed as RainDS-Real and RainDS-Syn, respectively. 
These datasets contain images of scenes that include rain streaks only (RS), raindrops only (RD), or both(RDS). 
RainDS-Syn has 3600 image pairs, with 3000 images used for training and the remaining 600 images for testing. 
Meanwhile, RainDS-Real comprises of 750 images, with 450 images for training and 300 images for evaluation.
\vspace{-0.2cm}
\subsection{Experimental Evaluation on Benchmarks}
\vspace{-0.1cm}
\noindent\textbf{Compared Methods.} Regarding removing raindrops and rain streaks in images, we conduct extensive experiments to compare various algorithms that can be used for image rain removal. (i) We compare previous SOTA methods (including GMM~\cite{GMM}, JCAS~\cite{JCAS}, DDN~\cite{DDN}, NLEDN~\cite{NLEDN}, RESCAN~\cite{rescan}, PreNet~\cite{prenet}, UMRL~\cite{UMRL}, JORDER-E~\cite{JORDER-E}, MSPFN~\cite{mspfn}, CCN~\cite{quaninonego}, IDT~\cite{IDT}). 
(ii) We also compare our network against universal image restoration methods, including MPRNet~\cite{mpr}, DGUNet~\cite{mou2022deep}, Uformer~\cite{wang2022uformer}, Restormer~\cite{zamir2021restormer}, NAFNet~\cite{chen2022simple}). 
For the specific raindrop removal dataset, we compare against Eigen’s model~\cite{eigen2013restoring}, Pix2Pix~\cite{pix2pix}, AttentGAN~\cite{qian2018attengan}, Quan’s network~\cite{quan2019deep}, CCN~\cite{quaninonego}, and IDT~\cite{IDT}. 
In the absence of pre-trained models, we conduct model retraining by utilizing publicly available code and subsequently evaluate the best model performance on test datasets to ensure a fair comparison.

\begin{table}[!t]
\setlength{\abovecaptionskip}{0cm} %调整caption与图的距离
\setlength{\belowcaptionskip}{-0.3cm}%调整caption与下文的距离
\centering
\caption{\small{Quantitative results compared with SOTA methods on the AGAN-Data~\cite{qian2018attengan}. Red and blue indicate the first and second best results.}}\label{raindroprealesults}
\resizebox{5.5cm}{!}{
\renewcommand\arraystretch{1.1}

\begin{tabular}{l||cccccc}
\hline\thickhline
\rowcolor{mygray}
{Method} & \multicolumn{2}{c}{PSNR$\uparrow$} & \multicolumn{2}{c}{SSIM$\uparrow$}\\
\hline\hline 
(ICCV'2013)Eigen’s model~\cite{eigen2013restoring} & \multicolumn{2}{c}{21.31}  & \multicolumn{2}{c}{0.757}\\
(CVPR2017)Pix2Pix~\cite{pix2pix} & \multicolumn{2}{c}{27.20}  &\multicolumn{2}{c}{0.836} \\
(CVPR'2018)AttenGAN~\cite{qian2018attengan} &  \multicolumn{2}{c}{${{31.59}}$} & \multicolumn{2}{c}{0.917}\\
(ICCV'2019)Quan’s network~\cite{quan2019deep} & \multicolumn{2}{c}{31.37} & \multicolumn{2}{c}{0.918} \\
(CVPR'2021)CCN~\cite{quaninonego} & \multicolumn{2}{c}{31.34} & \multicolumn{2}{c}{0.929} \\
(TPAMI'2022)IDT~\cite{IDT} & \multicolumn{2}{c}{$\mathbf{\textcolor{blue}{31.63}}$} & \multicolumn{2}{c}{$\mathbf{\textcolor{blue}{0.936}}$} \\
\hline\hline  UDR-S$^{2}$Former & \multicolumn{2}{c}{$\mathbf{\textcolor{red}{32.64}}$}& \multicolumn{2}{c}{$\mathbf{\textcolor{red}{0.943}}$} \\
\hline\thickhline
\end{tabular}}
\vspace{-0.2cm}
\end{table}

\noindent\textbf{Quantitative Comparison.} Tables ~\ref{rainstreaksresults},~\ref{raindsresults}, and~\ref{raindroprealesults} reports the quantitative results of our network and state-of-the-art methods on four benchmark datasets, which are Rain200H, Rain200L, RainDS, and AGAN-Data. 
As demonstrated in these Tables, our approach delivers superior metric results over compared state-of-the-art (SOTA) methods for single rain streaks or raindrop degradations. 
Moreover, by incorporating uncertainty to steer the network's attention towards challenging degradations, our method outperforms the IDT algorithm by 0.49dB, 0.22dB and 1.01dB on the Rain200H~\cite{JORDER}, Rain200L~\cite{JORDER} datasets and AGAN-Data~\cite{qian2018attengan}, while significantly outperforming previous uncertainty-based~\cite{UMRL} methods. 
Table ~\ref{raindsresults} demonstrates that our approach surpasses all prior designs for jointly removing rain streaks and raindrops. While networks like Restormer~\cite{zamir2021restormer} and IDT~\cite{IDT} can model long-distance dependencies sufficiently, our method is able to effectively model complex degradation relationships in rain-dominated landscapes (i.e., raindrops and rain streaks both) with uncertainty. Additionally, sparse sampling further facilitates the network's ability to obtain crucial global degradation information at a low cost, see Table \ref{raindsresults}. Based on the indicators, our approach exhibits a significant advantage over the latest designs, including Restormer (36.08PSNR $\rightarrow$ 36.91PSNR) and IDT (36.23PSNR $\rightarrow$ 36.91PSNR), in both real-world and synthetic datasets. Additionally, we also present the comparison of speed and memory cost in the supplementary material to further demonstrate our superiority.

\begin{figure*}[!t]
    \centering
    \setlength{\abovecaptionskip}{0.03cm} %调整caption与图的距离
    \setlength{\belowcaptionskip}{-0.4cm}%调整caption与下文的距离
    \includegraphics[width=17.2cm]{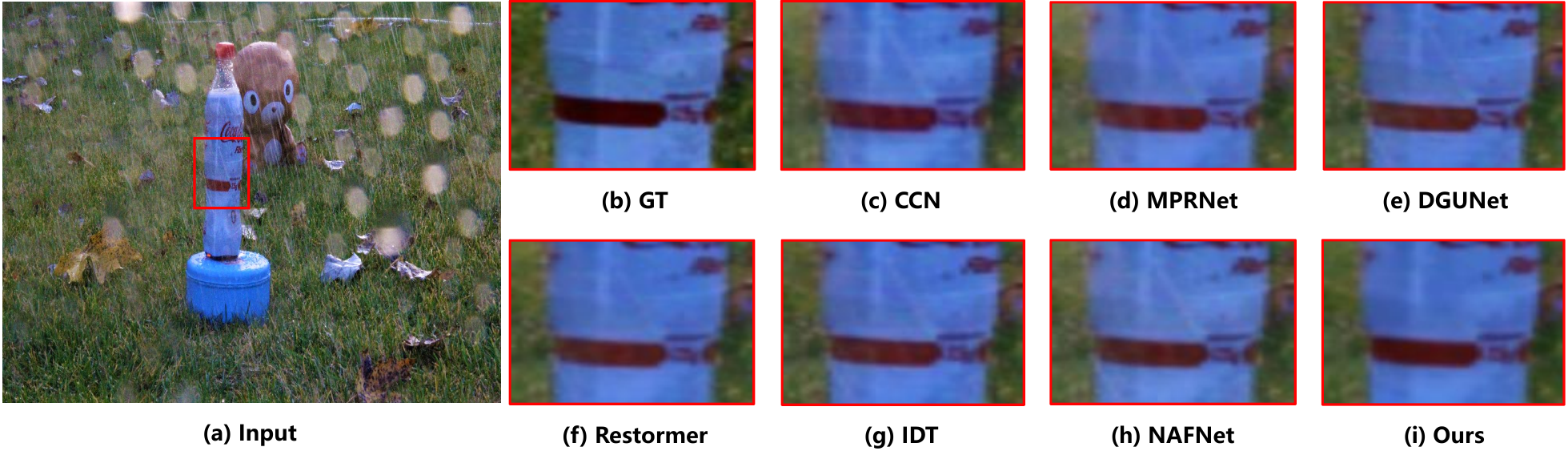}
    \caption{\small{Visual results for real-world sample of RainDS-Real~\cite{quaninonego} dataset}.}
    \label{real1}
\end{figure*}

\begin{figure}[!t]
    \centering
    \setlength{\abovecaptionskip}{0.05cm} %调整caption与图的距离
    \setlength{\belowcaptionskip}{-0.3cm}%调整caption与下文的距离
    \includegraphics[width=8.5cm]{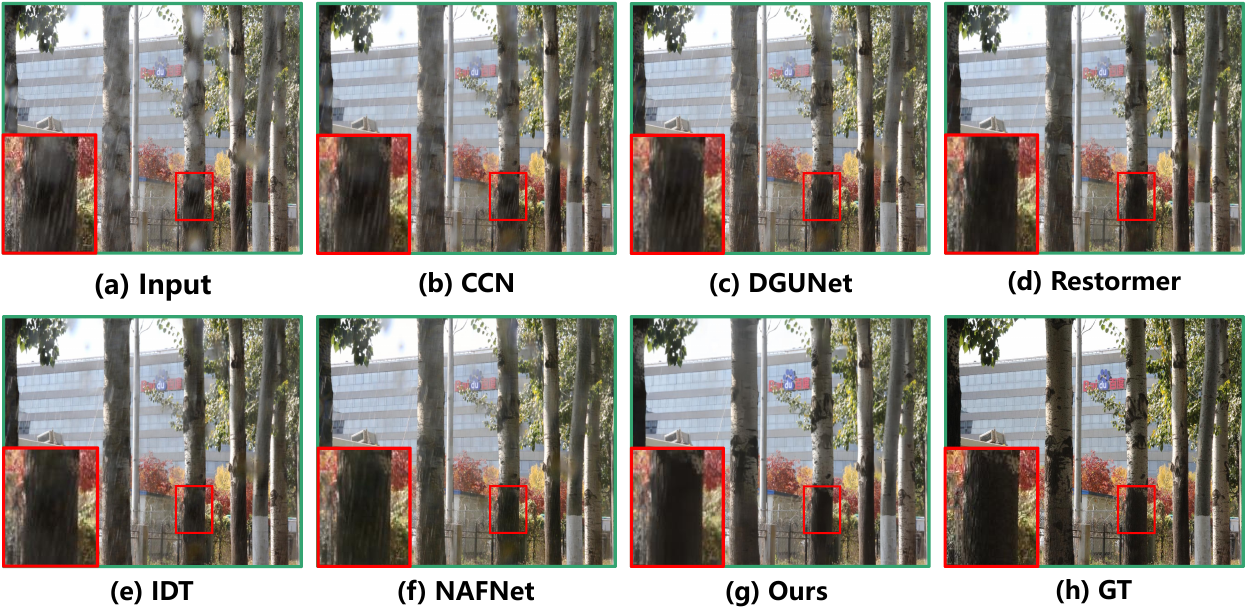}
    \caption{\small{Visual results for real-world sample of RainDS-Real~\cite{quaninonego} dataset}, Please zoom it for a better illustration.}
    \label{real}
\end{figure}

\noindent\textbf{Visual Comparison.} We present the comparisons of the visual effects in Fig.\ref{syn1}-\ref{real}. As depicted in the figures, we observe that in a scene containing rain streaks and raindrops, the scene information is either occluded by a broad range of raindrops or masked by dense rain streaks of varying scales. Other approaches either struggle to remove complex degradations or significantly compromise image details. Our approach leverages degradation relationship modeling to facilitate image restoration, and utilizes the clean information to recover intricate rain degradations. The resulting output exhibits higher fidelity with the more accurate restoration of great details. In real-world scenarios, our method enables the removal of various forms of degradation while recovering fine details, outperforming previous methods. 
%Please refer to our supplementary material for more visual comparisons.
\vspace{-0.2cm}
\section{Ablation Studies}
\vspace{-0.1cm}
To validate the effectiveness of our proposed design, we performed ablation experiments on UDR-S$^{2}$Former. To this end, we utilized the training dataset of RainDS-Syn~\cite{quaninonego} for training our model, and evaluated its performance on the RDS scenes of the corresponding testing set. The experimental settings are kept consistent with the aforementioned ones. In the subsequent sections, we analyze the individual contributions of each module toward the performance.

\vspace{-0.2cm}

\vspace{-0.1cm}
\subsection{Effectiveness of Image Reconstruction Module}
\begin{figure}[!t]
    \centering
    \setlength{\abovecaptionskip}{0.00cm} %调整caption与图的距离
    \setlength{\belowcaptionskip}{-0.4cm}%调整caption与下文的距离
    \includegraphics[width=8.5cm]{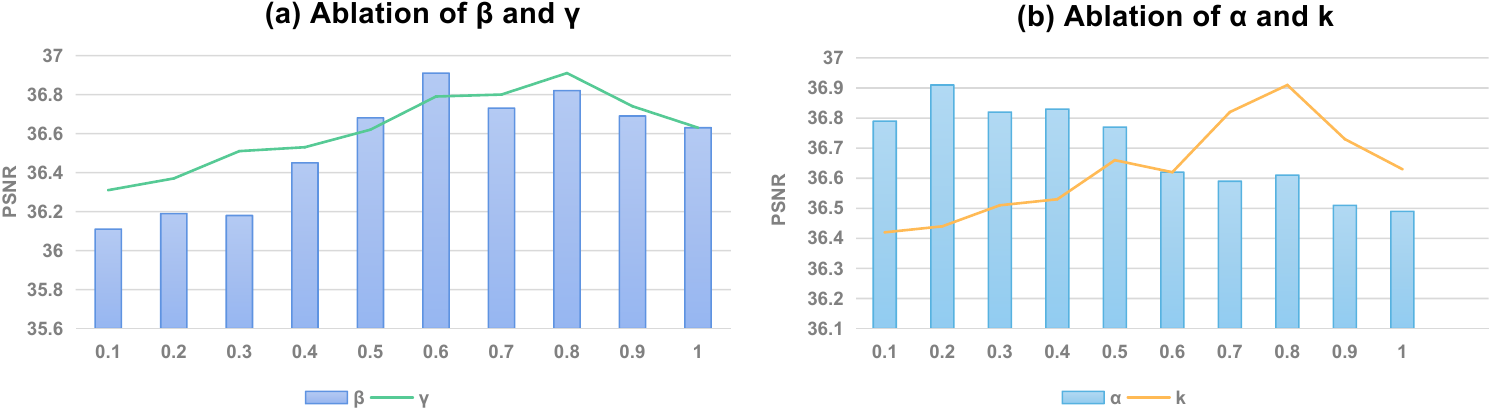}
    \caption{\small{About the ablation experiments of the parameter $\beta$, $\gamma$, $\alpha$ and \text{k}, the vertical axis is the PSNR metric.}}
    \label{parameter}
\end{figure}

\vspace{-0.1cm}
\noindent\textbf{Ablation researches of parameters mentioned in IRM.}
Our ablation studies involve thorough research of the parameters employed in the method, and we have depicted the results in Fig.\ref{parameter}. As demonstrated, the optimal value for each parameter lies within a particular range, which can be attributed to our initial design motivation holding theoretical significance within that range.

\noindent\textbf{Improvements of Sparse Sampling Attention (SSA) in IRM.}
We conducted a comparison of our sparse sampling attention with sparse attention in ART~\cite{ART} (SA), window-based self-attention~\cite{IDT} (WSA), and channel-dimension self-attention~\cite{zamir2021restormer} (CSA). The results are shown in Table \ref{ssa}. It demonstrates that sparse sampling attention performs significantly better than previous designs, underscoring the importance of incorporating degradation information by adaptive sampling from a global perspective for the unified removal of complex rain scenes.

\begin{table}[h]
\parbox{.24\textwidth}{
\centering
\setlength{\abovecaptionskip}{0cm} %调整caption与图的距离
\setlength{\belowcaptionskip}{-0.1cm}%调整caption与下文的距离
\caption{\small{Ablation studies on SSA (\S\ref{sec.ssa}). }}\label{ssa}
\resizebox{4cm}{!}{
\renewcommand\arraystretch{1.1}
\begin{tabular}{cccc}
\hline\thickhline
\rowcolor{mygray}
\textbf{Setting} & \textbf{Model}  & \textbf{PSNR} & \textbf{SSIM}  \\
\hline\thickhline
i & SA~\cite{ART}  & 36.39 & 0.962\\
ii& WSA~\cite{IDT}  & 36.15 & 0.959 \\
iii& CSA~\cite{zamir2021restormer}   & 36.23 & 0.960 \\
iv& SSA (Ours) & \underline{36.91} & \underline{0.966} \\
\hline\thickhline
\end{tabular}}
}
\hfill
\parbox{.23\textwidth}{
\centering
\setlength{\abovecaptionskip}{0cm} %调整caption与图的距离
\setlength{\belowcaptionskip}{-0.1cm}%调整caption与下文的距离
\caption{\small{Ablation studies on UDR. (\S\ref{sec.ssa} and \ref{sec.lr}). }}\label{UDR}
\resizebox{3.7cm}{!}{
\renewcommand\arraystretch{1.1}
\begin{tabular}{cccc}
\hline\thickhline
\rowcolor{mygray}
\textbf{Setting} & \textbf{Model}  & \textbf{PSNR} & \textbf{SSIM}  \\
\hline\thickhline
i & SSA w/o UD  & 36.25 & 0.960\\
ii & SSA w/o RS  & 36.29 & 0.961\\
iii& LR w/o UD  & 36.33 & 0.962 \\
iv& LR w/o RS & {36.45} & {0.963} \\
v& Ours & \underline{36.91} & \underline{0.966} \\
\hline\thickhline
\end{tabular}}
}
\end{table}

\noindent\textbf{Gains of Uncertainty-Driven Ranking (UDR). } In this section, we perform ablation studies of our uncertainty driven (UD) and ranking strategies (RS). The results in Table \ref{UDR} demonstrate that leveraging uncertainty can effectively constrain the process of sparse sampling and facilitate the network to recover locally degraded regions. However, using uncertainty maps directly without employing a ranking strategy leads to a significant drop in performance, due to the lack of using explicit properties of uncertainty maps.

%Please refer to our supplementary material for more ablation studies.
\vspace{-0.3cm}
\section{Conclusion}
\vspace{-0.1cm}
In this paper, we present a sparse sampling transformer with the uncertainty-driven ranking that removes rain streaks and raindrops in a unified approach. Our approach employs sparse sampling self-attention to effectively capture global degradation relationships in an adaptive manner. We explicitly leverage the uncertainty map via a ranking strategy to constrain the sampling process and facilitate local reconstruction. Our method achieves SOTA results while requiring minimal computational overhead, showcasing its superiority over existing approaches.

%\vspace{-4mm}
\paragraph{Acknowledgment.}
This work was supported by Guangzhou Municipal Science and Technology Project (Grant No. 2023A03J0671), and the National Natural Science Foundation of China (Grant No. 61902275).

{\small
\bibliographystyle{ieee_fullname}
\bibliography{egbib}

\begin{thebibliography}{10}\itemsep=-1pt

\bibitem{badrinarayanan2017segnet}
Vijay Badrinarayanan, Alex Kendall, and Roberto Cipolla.
\newblock Segnet: A deep convolutional encoder-decoder architecture for image
  segmentation.
\newblock {\em IEEE transactions on pattern analysis and machine intelligence},
  39(12):2481--2495, 2017.

\bibitem{chang2020data}
Jie Chang, Zhonghao Lan, Changmao Cheng, and Yichen Wei.
\newblock Data uncertainty learning in face recognition.
\newblock In {\em Proceedings of the IEEE/CVF conference on computer vision and
  pattern recognition}, pages 5710--5719, 2020.

\bibitem{chen2023masked}
Haoyu Chen, Jinjin Gu, Yihao Liu, Salma~Abdel Magid, Chao Dong, Qiong Wang,
  Hanspeter Pfister, and Lei Zhu.
\newblock Masked image training for generalizable deep image denoising.
\newblock In {\em Proceedings of the IEEE/CVF Conference on Computer Vision and
  Pattern Recognition}, pages 1692--1703, 2023.

\bibitem{chen2021attention}
Haoyu Chen, Jinjin Gu, and Zhi Zhang.
\newblock Attention in attention network for image super-resolution.
\newblock {\em arXiv preprint arXiv:2104.09497}, 2021.

\bibitem{chen2022simple}
Liangyu Chen, Xiaojie Chu, Xiangyu Zhang, and Jian Sun.
\newblock Simple baselines for image restoration.
\newblock In {\em European Conference on Computer Vision}, pages 17--33.
  Springer, 2022.

\bibitem{chen2021hinet}
Liangyu Chen, Xin Lu, Jie Zhang, Xiaojie Chu, and Chengpeng Chen.
\newblock Hinet: Half instance normalization network for image restoration.
\newblock In {\em Proceedings of the IEEE/CVF Conference on Computer Vision and
  Pattern Recognition}, pages 182--192, 2021.

\bibitem{chen2022dual}
Sixiang Chen, Tian Ye, Yun Liu, and Erkang Chen.
\newblock Dual-former: Hybrid self-attention transformer for efficient image
  restoration.
\newblock {\em arXiv preprint arXiv:2210.01069}, 2022.

\bibitem{chen2022snowformer}
Sixiang Chen, Tian Ye, Yun Liu, Erkang Chen, Jun Shi, and Jingchun Zhou.
\newblock Snowformer: Scale-aware transformer via context interaction for
  single image desnowing.
\newblock {\em arXiv preprint arXiv:2208.09703}, 2022.

\bibitem{chen2022msp}
Sixiang Chen, Tian Ye, Yun Liu, Taodong Liao, Yi Ye, and Erkang Chen.
\newblock Msp-former: Multi-scale projection transformer for single image
  desnowing.
\newblock {\em arXiv preprint arXiv:2207.05621}, 2022.

\bibitem{chen2023dehrformer}
Sixiang Chen, Tian Ye, Jun Shi, Yun Liu, JingXia Jiang, Erkang Chen, and Peng
  Chen.
\newblock Dehrformer: Real-time transformer for depth estimation and haze
  removal from varicolored haze scenes.
\newblock In {\em ICASSP 2023-2023 IEEE International Conference on Acoustics,
  Speech and Signal Processing (ICASSP)}, pages 1--5. IEEE, 2023.

\bibitem{chen2023learning}
Xiang Chen, Hao Li, Mingqiang Li, and Jinshan Pan.
\newblock Learning a sparse transformer network for effective image deraining.
\newblock In {\em Proceedings of the IEEE/CVF Conference on Computer Vision and
  Pattern Recognition}, pages 5896--5905, 2023.

\bibitem{DRD}
Sen Deng, Mingqiang Wei, Jun Wang, Yidan Feng, Luming Liang, Haoran Xie, Fu~Lee
  Wang, and Meng Wang.
\newblock Detail-recovery image deraining via context aggregation networks.
\newblock In {\em Proceedings of the IEEE/CVF conference on computer vision and
  pattern recognition}, pages 14560--14569, 2020.

\bibitem{vit}
Alexey Dosovitskiy, Lucas Beyer, Alexander Kolesnikov, Dirk Weissenborn,
  Xiaohua Zhai, Thomas Unterthiner, Mostafa Dehghani, Matthias Minderer, Georg
  Heigold, Sylvain Gelly, et~al.
\newblock An image is worth 16x16 words: Transformers for image recognition at
  scale.
\newblock {\em arXiv preprint arXiv:2010.11929}, 2020.

\bibitem{eigen2013restoring}
David Eigen, Dilip Krishnan, and Rob Fergus.
\newblock Restoring an image taken through a window covered with dirt or rain.
\newblock In {\em Proceedings of the IEEE international conference on computer
  vision}, pages 633--640, 2013.

\bibitem{DDN}
Xueyang Fu, Jiabin Huang, Delu Zeng, Yue Huang, Xinghao Ding, and John Paisley.
\newblock Removing rain from single images via a deep detail network.
\newblock In {\em Proceedings of the IEEE conference on computer vision and
  pattern recognition}, pages 3855--3863, 2017.

\bibitem{gu2020image}
Jinjin Gu, Haoming Cai, Haoyu Chen, Xiaoxing Ye, Jimmy Ren, and Chao Dong.
\newblock Image quality assessment for perceptual image restoration: A new
  dataset, benchmark and metric.
\newblock {\em arXiv preprint arXiv:2011.15002}, 2020.

\bibitem{gu2020pipal}
Jinjin Gu, Haoming Cai, Haoyu Chen, Xiaoxing Ye, Jimmy Ren, and Chao Dong.
\newblock Pipal: a large-scale image quality assessment dataset for perceptual
  image restoration.
\newblock In {\em European Conference on Computer Vision}, pages 633--651.
  Springer, 2020.

\bibitem{JCAS}
Shuhang Gu, Deyu Meng, Wangmeng Zuo, and Lei Zhang.
\newblock Joint convolutional analysis and synthesis sparse representation for
  single image layer separation.
\newblock In {\em Proceedings of the IEEE International Conference on Computer
  Vision}, pages 1708--1716, 2017.

\bibitem{hong2022uncertainty}
Ming Hong, Jianzhuang Liu, Cuihua Li, and Yanyun Qu.
\newblock Uncertainty-driven dehazing network.
\newblock In {\em Proceedings of the AAAI Conference on Artificial
  Intelligence}, volume~36, pages 906--913, 2022.

\bibitem{huang2022deep}
Jie Huang, Yajing Liu, Feng Zhao, Keyu Yan, Jinghao Zhang, Yukun Huang, Man
  Zhou, and Zhiwei Xiong.
\newblock Deep fourier-based exposure correction network with spatial-frequency
  interaction.
\newblock In {\em European Conference on Computer Vision}, pages 163--180.
  Springer, 2022.

\bibitem{HPEU}
Jie Huang, Zhiwei Xiong, Xueyang Fu, Dong Liu, and Zheng-Jun Zha.
\newblock Hybrid image enhancement with progressive laplacian enhancing unit.
\newblock In {\em Proceedings of the 27th ACM International Conference on
  Multimedia}, page 1614–1622, 2019.

\bibitem{Huang_2023_CVPR}
Jie Huang, Feng Zhao, Man Zhou, Jie Xiao, Naishan Zheng, Kaiwen Zheng, and
  Zhiwei Xiong.
\newblock Learning sample relationship for exposure correction.
\newblock In {\em Proceedings of the IEEE/CVF Conference on Computer Vision and
  Pattern Recognition (CVPR)}, pages 9904--9913, June 2023.

\bibitem{ECLNet}
Jie Huang, Man Zhou, Yajing Liu, Mingde Yao, Feng Zhao, and Zhiwei Xiong.
\newblock Exposure-consistency representation learning for exposure correction.
\newblock In {\em Proceedings of the 30th ACM International Conference on
  Multimedia}, page 6309–6317, 2022.

\bibitem{Huang_2018_ECCV_Workshops}
Jie Huang, Pengfei Zhu, Mingrui Geng, Jiewen Ran, Xingguang Zhou, Chen Xing,
  Pengfei Wan, and Xiangyang Ji.
\newblock Range scaling global u-net for perceptual image enhancement on mobile
  devices.
\newblock In {\em Proceedings of the European Conference on Computer Vision
  (ECCV) Workshops}, September 2018.

\bibitem{PSNR}
Quan Huynh-Thu and Mohammed Ghanbari.
\newblock Scope of validity of psnr in image/video quality assessment.
\newblock {\em Electronics letters}, 44(13):800--801, 2008.

\bibitem{pix2pix}
Phillip Isola, Jun-Yan Zhu, Tinghui Zhou, and Alexei~A Efros.
\newblock Image-to-image translation with conditional adversarial networks.
\newblock In {\em Proceedings of the IEEE conference on computer vision and
  pattern recognition}, pages 1125--1134, 2017.

\bibitem{jiang2022magic}
Kui Jiang, Zhongyuan Wang, Chen Chen, Zheng Wang, Laizhong Cui, and Chia-Wen
  Lin.
\newblock Magic elf: Image deraining meets association learning and
  transformer.
\newblock {\em arXiv preprint arXiv:2207.10455}, 2022.

\bibitem{mspfn}
Kui Jiang, Zhongyuan Wang, Peng Yi, Chen Chen, Baojin Huang, Yimin Luo, Jiayi
  Ma, and Junjun Jiang.
\newblock Multi-scale progressive fusion network for single image deraining.
\newblock In {\em Proceedings of the IEEE/CVF conference on computer vision and
  pattern recognition}, pages 8346--8355, 2020.

\bibitem{jin2023estimating}
Yeying Jin, Ruoteng Li, Wenhan Yang, and Robby~T Tan.
\newblock Estimating reflectance layer from a single image: Integrating
  reflectance guidance and shadow/specular aware learning.
\newblock In {\em Proceedings of the AAAI Conference on Artificial
  Intelligence}, volume~37, pages 1069--1077, 2023.

\bibitem{jin2023enhancing}
Yeying Jin, Beibei Lin, Wending Yan, Wei Ye, Yuan Yuan, and Robby~T. Tan.
\newblock Enhancing visibility in nighttime haze images using guided apsf and
  gradient adaptive convolution, 2023.

\bibitem{jin2021dc}
Yeying Jin, Aashish Sharma, and Robby~T Tan.
\newblock Dc-shadownet: Single-image hard and soft shadow removal using
  unsupervised domain-classifier guided network.
\newblock In {\em Proceedings of the IEEE/CVF International Conference on
  Computer Vision}, pages 5027--5036, 2021.

\bibitem{jin2022structure}
Yeying Jin, Wending Yan, Wenhan Yang, and Robby~T Tan.
\newblock Structure representation network and uncertainty feedback learning
  for dense non-uniform fog removal.
\newblock In {\em Proceedings of the Asian Conference on Computer Vision},
  pages 2041--2058, 2022.

\bibitem{jin2022unsupervised}
Yeying Jin, Wenhan Yang, and Robby~T Tan.
\newblock Unsupervised night image enhancement: When layer decomposition meets
  light-effects suppression.
\newblock In {\em European Conference on Computer Vision}, pages 404--421.
  Springer, 2022.

\bibitem{kendall2017uncertainties}
Alex Kendall and Yarin Gal.
\newblock What uncertainties do we need in bayesian deep learning for computer
  vision?
\newblock {\em Advances in neural information processing systems}, 30, 2017.

\bibitem{NLEDN}
Guanbin Li, Xiang He, Wei Zhang, Huiyou Chang, Le Dong, and Liang Lin.
\newblock Non-locally enhanced encoder-decoder network for single image
  de-raining.
\newblock In {\em Proceedings of the 26th ACM international conference on
  Multimedia}, pages 1056--1064, 2018.

\bibitem{li2019heavy}
Ruoteng Li, Loong-Fah Cheong, and Robby~T Tan.
\newblock Heavy rain image restoration: Integrating physics model and
  conditional adversarial learning.
\newblock In {\em Proceedings of the IEEE/CVF conference on computer vision and
  pattern recognition}, pages 1633--1642, 2019.

\bibitem{rescan}
Xia Li, Jianlong Wu, Zhouchen Lin, Hong Liu, and Hongbin Zha.
\newblock Recurrent squeeze-and-excitation context aggregation net for single
  image deraining.
\newblock In {\em Proceedings of the European conference on computer vision
  (ECCV)}, pages 254--269, 2018.

\bibitem{GMM}
Yu Li, Robby~T Tan, Xiaojie Guo, Jiangbo Lu, and Michael~S Brown.
\newblock Rain streak removal using layer priors.
\newblock In {\em Proceedings of the IEEE conference on computer vision and
  pattern recognition}, pages 2736--2744, 2016.

\bibitem{liang2021swinir}
Jingyun Liang, Jiezhang Cao, Guolei Sun, Kai Zhang, Luc Van~Gool, and Radu
  Timofte.
\newblock Swinir: Image restoration using swin transformer.
\newblock In {\em Proceedings of the IEEE/CVF International Conference on
  Computer Vision}, pages 1833--1844, 2021.

\bibitem{liang2022drt}
Yuanchu Liang, Saeed Anwar, and Yang Liu.
\newblock Drt: A lightweight single image deraining recursive transformer.
\newblock In {\em Proceedings of the IEEE/CVF Conference on Computer Vision and
  Pattern Recognition}, pages 589--598, 2022.

\bibitem{liu2023nighthazeformer}
Yun Liu, Zhongsheng Yan, Sixiang Chen, Tian Ye, Wenqi Ren, and Erkang Chen.
\newblock Nighthazeformer: Single nighttime haze removal using prior query
  transformer.
\newblock {\em arXiv preprint arXiv:2305.09533}, 2023.

\bibitem{swim}
Ze Liu, Yutong Lin, Yue Cao, Han Hu, Yixuan Wei, Zheng Zhang, Stephen Lin, and
  Baining Guo.
\newblock Swin transformer: Hierarchical vision transformer using shifted
  windows.
\newblock In {\em Proceedings of the IEEE/CVF international conference on
  computer vision}, pages 10012--10022, 2021.

\bibitem{mou2022deep}
Chong Mou, Qian Wang, and Jian Zhang.
\newblock Deep generalized unfolding networks for image restoration.
\newblock In {\em Proceedings of the IEEE/CVF Conference on Computer Vision and
  Pattern Recognition}, pages 17399--17410, 2022.

\bibitem{ning2021uncertainty}
Qian Ning, Weisheng Dong, Xin Li, Jinjian Wu, and Guangming Shi.
\newblock Uncertainty-driven loss for single image super-resolution.
\newblock {\em Advances in Neural Information Processing Systems},
  34:16398--16409, 2021.

\bibitem{paszke2019pytorch}
Adam Paszke, Sam Gross, Francisco Massa, Adam Lerer, James Bradbury, Gregory
  Chanan, Trevor Killeen, Zeming Lin, Natalia Gimelshein, Luca Antiga, et~al.
\newblock Pytorch: An imperative style, high-performance deep learning library.
\newblock {\em Advances in neural information processing systems}, 32, 2019.

\bibitem{purohit2021spatially}
Kuldeep Purohit, Maitreya Suin, AN Rajagopalan, and Vishnu~Naresh Boddeti.
\newblock Spatially-adaptive image restoration using distortion-guided
  networks.
\newblock In {\em Proceedings of the IEEE/CVF International Conference on
  Computer Vision}, pages 2309--2319, 2021.

\bibitem{qian2018attengan}
Rui Qian, Robby~T Tan, Wenhan Yang, Jiajun Su, and Jiaying Liu.
\newblock Attentive generative adversarial network for raindrop removal from a
  single image.
\newblock In {\em Proceedings of the IEEE conference on computer vision and
  pattern recognition}, pages 2482--2491, 2018.

\bibitem{quaninonego}
Ruijie Quan, Xin Yu, Yuanzhi Liang, and Yi Yang.
\newblock Removing raindrops and rain streaks in one go.
\newblock In {\em Proceedings of the IEEE/CVF Conference on Computer Vision and
  Pattern Recognition}, pages 9147--9156, 2021.

\bibitem{quan2019deep}
Yuhui Quan, Shijie Deng, Yixin Chen, and Hui Ji.
\newblock Deep learning for seeing through window with raindrops.
\newblock In {\em Proceedings of the IEEE/CVF International Conference on
  Computer Vision}, pages 2463--2471, 2019.

\bibitem{prenet}
Dongwei Ren, Wangmeng Zuo, Qinghua Hu, Pengfei Zhu, and Deyu Meng.
\newblock Progressive image deraining networks: A better and simpler baseline.
\newblock In {\em Proceedings of the IEEE/CVF Conference on Computer Vision and
  Pattern Recognition}, pages 3937--3946, 2019.

\bibitem{simonyan2014very}
Karen Simonyan and Andrew Zisserman.
\newblock Very deep convolutional networks for large-scale image recognition.
\newblock {\em arXiv preprint arXiv:1409.1556}, 2014.

\bibitem{song2022vision}
Yuda Song, Zhuqing He, Hui Qian, and Xin Du.
\newblock Vision transformers for single image dehazing.
\newblock {\em arXiv preprint arXiv:2204.03883}, 2022.

\bibitem{tong2022semi}
Ming Tong, Yongzhen Wang, Peng Cui, Xuefeng Yan, and Mingqiang Wei.
\newblock Semi-uformer: Semi-supervised uncertainty-aware transformer for image
  dehazing.
\newblock {\em arXiv preprint arXiv:2210.16057}, 2022.

\bibitem{valanarasu2022transweather}
Jeya Maria~Jose Valanarasu, Rajeev Yasarla, and Vishal~M Patel.
\newblock Transweather: Transformer-based restoration of images degraded by
  adverse weather conditions.
\newblock In {\em Proceedings of the IEEE/CVF Conference on Computer Vision and
  Pattern Recognition}, pages 2353--2363, 2022.

\bibitem{wang2019spatial}
Tianyu Wang, Xin Yang, Ke Xu, Shaozhe Chen, Qiang Zhang, and Rynson~WH Lau.
\newblock Spatial attentive single-image deraining with a high quality real
  rain dataset.
\newblock In {\em Proceedings of the IEEE/CVF Conference on Computer Vision and
  Pattern Recognition}, pages 12270--12279, 2019.

\bibitem{SSIM}
Zhou Wang, Alan~C Bovik, Hamid~R Sheikh, and Eero~P Simoncelli.
\newblock Image quality assessment: from error visibility to structural
  similarity.
\newblock {\em IEEE transactions on image processing}, 13(4):600--612, 2004.

\bibitem{wang2022uformer}
Zhendong Wang, Xiaodong Cun, Jianmin Bao, Wengang Zhou, Jianzhuang Liu, and
  Houqiang Li.
\newblock Uformer: A general u-shaped transformer for image restoration.
\newblock In {\em Proceedings of the IEEE/CVF Conference on Computer Vision and
  Pattern Recognition}, pages 17683--17693, 2022.

\bibitem{IDT}
Jie Xiao, Xueyang Fu, Aiping Liu, Feng Wu, and Zheng-Jun Zha.
\newblock Image de-raining transformer.
\newblock {\em IEEE Transactions on Pattern Analysis and Machine Intelligence},
  2022.

\bibitem{xu2021intensity}
Ke Xu, Xin Tian, Xin Yang, Baocai Yin, and Rynson~WH Lau.
\newblock Intensity-aware single-image deraining with semantic and color
  regularization.
\newblock {\em IEEE Transactions on Image Processing}, 30:8497--8509, 2021.

\bibitem{JORDER-E}
Wenhan Yang, Robby~T Tan, Jiashi Feng, Zongming Guo, Shuicheng Yan, and Jiaying
  Liu.
\newblock Joint rain detection and removal from a single image with
  contextualized deep networks.
\newblock {\em IEEE transactions on pattern analysis and machine intelligence},
  42(6):1377--1393, 2019.

\bibitem{JORDER}
Wenhan Yang, Robby~T Tan, Jiashi Feng, Jiaying Liu, Zongming Guo, and Shuicheng
  Yan.
\newblock Deep joint rain detection and removal from a single image.
\newblock In {\em Proceedings of the IEEE conference on computer vision and
  pattern recognition}, pages 1357--1366, 2017.

\bibitem{yang2021recurrent}
Wenhan Yang, Robby~T Tan, Jiashi Feng, Shiqi Wang, Bin Cheng, and Jiaying Liu.
\newblock Recurrent multi-frame deraining: Combining physics guidance and
  adversarial learning.
\newblock {\em IEEE Transactions on Pattern Analysis and Machine Intelligence},
  44(11):8569--8586, 2021.

\bibitem{UMRL}
Rajeev Yasarla and Vishal~M Patel.
\newblock Uncertainty guided multi-scale residual learning-using a cycle
  spinning cnn for single image de-raining.
\newblock In {\em Proceedings of the IEEE/CVF Conference on Computer Vision and
  Pattern Recognition}, pages 8405--8414, 2019.

\bibitem{ye2021perceiving}
Tian Ye, Yunchen Zhang, Mingchao Jiang, Liang Chen, Yun Liu, Sixiang Chen, and
  Erkang Chen.
\newblock Perceiving and modeling density for image dehazing.
\newblock In {\em European Conference on Computer Vision}, pages 130--145.
  Springer, 2022.

\bibitem{you2013adherent}
Shaodi You, Robby~T Tan, Rei Kawakami, and Katsushi Ikeuchi.
\newblock Adherent raindrop detection and removal in video.
\newblock In {\em Proceedings of the IEEE Conference on Computer Vision and
  Pattern Recognition}, pages 1035--1042, 2013.

\bibitem{DehazeYu}
Hu Yu, Jie Huang, Yajing Liu, Qi Zhu, Man Zhou, and Feng Zhao.
\newblock Source-free domain adaptation for real-world image dehazing.
\newblock In {\em Proceedings of the 30th ACM International Conference on
  Multimedia}, page 6645–6654, 2022.

\bibitem{DehazeMMYu}
Hu Yu, Jie Huang, Yajing Liu, Qi Zhu, Man Zhou, and Feng Zhao.
\newblock Source-free domain adaptation for real-world image dehazing.
\newblock In {\em Proceedings of the 30th ACM International Conference on
  Multimedia}, page 6645–6654, 2022.

\bibitem{zamir2021restormer}
Syed~Waqas Zamir, Aditya Arora, Salman Khan, Munawar Hayat, Fahad~Shahbaz Khan,
  and Ming-Hsuan Yang.
\newblock Restormer: Efficient transformer for high-resolution image
  restoration.
\newblock In {\em Proceedings of the IEEE/CVF conference on computer vision and
  pattern recognition}, pages 5728--5739, 2022.

\bibitem{mpr}
Syed~Waqas Zamir, Aditya Arora, Salman Khan, Munawar Hayat, Fahad~Shahbaz Khan,
  Ming-Hsuan Yang, and Ling Shao.
\newblock Multi-stage progressive image restoration.
\newblock In {\em Proceedings of the IEEE/CVF Conference on Computer Vision and
  Pattern Recognition}, pages 14821--14831, 2021.

\bibitem{zhang2022dino}
Hao Zhang, Feng Li, Shilong Liu, Lei Zhang, Hang Su, Jun Zhu, Lionel Ni, and
  Harry Shum.
\newblock Dino: Detr with improved denoising anchor boxes for end-to-end object
  detection.
\newblock In {\em International Conference on Learning Representations}, 2022.

\bibitem{ART}
Jiale Zhang, Yulun Zhang, Jinjin Gu, Yongbing Zhang, Linghe Kong, and Xin Yuan.
\newblock Accurate image restoration with attention retractable transformer.
\newblock {\em arXiv preprint arXiv:2210.01427}, 2022.

\bibitem{zhang2018ffdnet}
Kai Zhang, Wangmeng Zuo, and Lei Zhang.
\newblock Ffdnet: Toward a fast and flexible solution for cnn-based image
  denoising.
\newblock {\em IEEE Transactions on Image Processing}, 27(9):4608--4622, 2018.

\bibitem{NEURIPS2022_91a23b3e}
man zhou, Hu Yu, Jie Huang, Feng Zhao, Jinwei Gu, Chen~Change Loy, Deyu Meng,
  and Chongyi Li.
\newblock Deep fourier up-sampling.
\newblock In {\em Advances in Neural Information Processing Systems},
  volume~35, pages 22995--23008. Curran Associates, Inc., 2022.

\bibitem{zhu2020learning}
Lei Zhu, Zijun Deng, Xiaowei Hu, Haoran Xie, Xuemiao Xu, Jing Qin, and
  Pheng-Ann Heng.
\newblock Learning gated non-local residual for single-image rain streak
  removal.
\newblock {\em IEEE Transactions on Circuits and Systems for Video Technology},
  31(6):2147--2159, 2020.

\bibitem{zhujoint}
Lei Zhu, Chi-Wing Fu, Dani Lischinski, and Pheng-Ann Heng.
\newblock Joint bi-layer optimization for single-image rain streak removal.
\newblock In {\em Proceedings of the IEEE international conference on computer
  vision}, pages 2526--2534, 2017.

\end{thebibliography}
}

\end{document}